# Relational Dynamic Bayesian Networks

**Sumit Sanghai**                                    SANGHAI@CS.WASHINGTON.EDU
**Pedro Domingos**                                   PEDROD@CS.WASHINGTON.EDU
**Daniel Weld**                                      WELD@CS.WASHINGTON.EDU
*Department of Computer Science and Engineering*
*University of Washington*

## Abstract

Stochastic processes that involve the creation of objects and relations over time are widespread, but relatively poorly studied. For example, accurate fault diagnosis in factory assembly processes requires inferring the probabilities of erroneous assembly operations, but doing this efficiently and accurately is difficult. Modeled as dynamic Bayesian networks, these processes have discrete variables with very large domains and extremely high dimensionality. In this paper, we introduce relational dynamic Bayesian networks (RDBNs), which are an extension of dynamic Bayesian networks (DBNs) to first-order logic. RDBNs are a generalization of dynamic probabilistic relational models (DPRMs), which we had proposed in our previous work to model dynamic uncertain domains. We first extend the Rao-Blackwellised particle filtering described in our earlier work to RDBNs. Next, we lift the assumptions associated with Rao-Blackwellization in RDBNs and propose two new forms of particle filtering. The first one uses abstraction hierarchies over the predicates to smooth the particle filter's estimates. The second employs kernel density estimation with a kernel function specifically designed for relational domains. Experiments show these two methods greatly outperform standard particle filtering on the task of assembly plan execution monitoring.

## 1. Introduction

Sequential phenomena abound in the world, and uncertainty is a common feature of them. Dynamic Bayesian networks (DBNs), one of the most powerful representations available for such phenomena, represent the state of the world as a set of variables, and model the probabilistic dependencies of the variables within and between time steps (Dean & Kanazawa, 1989). While a major advance over previous approaches, DBNs are essentially propositional, with no notion of objects or relations; hence DBNs are unable to compactly represent many real-world domains. For example, manufacturing plants assemble complex artifacts (*e.g.*, cars, computers, aircraft) from large numbers of component parts, using multiple kinds of machines and operations. Capturing such a domain in a DBN would require exhaustively representing all possible objects and relations among them, which is impractical.

Formalisms that can represent objects and relations, as opposed to just variables, have a long history in AI. Recently, significant progress has been made in combining them with a principled treatment of uncertainty. In particular, probabilistic relational models or PRMs (Friedman, Getoor, Koller, & Pfeffer, 1999) are an extension of Bayesian networks that allows reasoning with classes, objects and relations. Recently, we proposed dynamic probabilistic relational models (DPRMs) (Sanghai, Domingos, & Weld, 2003), which combine PRMs and DBNs to allow reasoning with classes, objects and relations in a dynamic environment. We also developed a relational Rao-Blackwellized particle filtering mechanism for state monitoring in DPRMs.

In this paper we introduce *relational dynamic Bayesian networks* (RDBNs) which extend DBNs to first-order (relational) domains. RDBNs subsume DPRMs and have several advantages over them, including greater simplicity and expressivity. Furthermore, they may be more easily learned using ILP techniques.

We develop a series of efficient inference procedures for RDBNs (which are also applicable to DPRMs or any other relational stochastic process model). The Rao-Blackwellised particle filtering described in our previous paper requires two strong assumptions which restrict its applicability. We lift these assumptions, developing two new forms of particle filtering. In the first approach, we build an abstraction hierarchy over the first-order predicates and use it to smooth the particle filter estimates. In the second approach, we introduce a variant of kernel density estimation with a kernel function specifically designed for relational domains.

Early fault detection can greatly reduce the cost of manufacturing processes. In this paper we apply our inference algorithms to execution-monitoring of assembly plans, showing that our methods scale to problems with over a thousand objects and thousands of steps. Other domains where our techniques may be helpful include robot control, vision in motion, language processing, computational modeling of markets, battlefield management, cell biology, ecosystem modeling, and analysis of Web information. The following are the significant contributions of this paper:

- We define *relational dynamic Bayesian networks* (RDBNs), which allow modeling uncertainty in dynamic relational domains.

- We present several novel methods for inferencing in RDBNs which use Rao-Blackwellization particle filtering, smoothing on relational abstraction hierarchies and relational kernel density estimation.

- We apply RDBNs to fault diagnosis in factory assembly processes, showing that the inference algorithms we propose outperform traditional particle filtering.

The rest of the paper is structured as follows. In Section 2 we review DBNs and briefly discuss the different filtering algorithms applicable to them. We introduce RDBNs in Section 3, and in Sections 4, 5 and 6 we describe our inference methods. In Section 7 we report our experimental results. In Section 8 we discuss the related work and we show that RDBNs subsume DPRMs. We conclude with a discussion of future work.

## 2. Background

A *Bayesian network* encodes the joint probability distribution of a set of variables, $\{Z_1, \ldots, Z_d\}$, as a directed acyclic graph and a set of conditional probability models. Each node corresponds to a variable, and the model associated with it allows us to compute the probability of a state of the variable given the state of its parents. The set of parents of $Z_i$, denoted $Pa(Z_i)$, is the set of nodes with an arc to $Z_i$ in the graph. The structure of the network encodes the assertion that each node is conditionally independent of its non-descendants given its parents. The probability of an arbitrary event $Z = (Z_1, \ldots, Z_d)$ can then be computed as $P(Z) = \prod_{i=1}^{d} P(Z_i | Pa(Z_i))$.

### 2.1 Dynamic Bayesian Networks

*Dynamic Bayesian Networks (DBNs)* (Dean & Kanazawa, 1989) are an extension of Bayesian networks for modeling dynamic systems. In a DBN, the state at time $t$ is represented by a set of random variables $Z_t = (Z_{1,t}, \ldots, Z_{d,t})$. The state at time $t$ is dependent on the states at previous time steps. Typically, we assume that each state only depends on the immediately preceding state (i.e., the system is first-order Markov), and thus we need to represent the transition distribution $P(Z_{t+1}|Z_t)$. This can be done using a two-time-slice Bayesian network fragment (2TBN) $B_{t+1}$, which contains variables from $Z_{t+1}$ whose parents are variables from $Z_t$ and/or $Z_{t+1}$, and variables from $Z_t$ without their parents. Typically, we also assume that the process is stationary, i.e., the transition models for all time slices are identical: $B_1 = B_2 = \ldots = B_t = B_{\rightarrow}$. Thus a DBN is defined to be a pair of Bayesian networks $(B_0, B_{\rightarrow})$, where $B_0$ represents the initial distribution $P(Z_0)$, and $B_{\rightarrow}$ is a two-time-slice Bayesian network, which as discussed above defines the transition distribution $P(Z_{t+1}|Z_t)$.

The set $Z_t$ is commonly divided into two sets: the unobserved state variables $X_t$ and the observed variables $Y_t$. The observed variables $Y_t$ are assumed to depend only on the current state variables $X_t$. The joint distribution represented by a DBN can then be obtained by unrolling the 2TBN:

$$P(X_0, ..., X_T, Y_0, ..., Y_T) = P(X_0)P(Y_0|X_0)\prod_{t=1}^{T} P(X_t|X_{t-1})P(Y_t|X_t) \qquad (1)$$

## 2.2 Inference in DBNs

Various types of inference are possible in DBNs. In this paper, we will focus on state monitoring (also known as filtering or tracking). However, the methods that we will propose later can be used in other types of inference.

The goal in state monitoring is to estimate the current state of the world given the observations made up to the present, i.e., to compute the distribution $P(X_T|Y_0, Y_1, ..., Y_T)$. Proper state monitoring is a necessary precondition for rational decision-making in dynamic domains. Since inference in DBNs is NP-complete, we usually resort to approximate methods, of which the most widely used one is *particle filtering* (Doucet, de Freitas, & Gordon, 2001). Particle filtering is a stochastic algorithm which maintains a set of particles (samples) $x_t^1, x_t^2, \ldots, x_t^N$ to approximately represent the distribution of possible states at time $t$ given the observations. Each particle $x_t^i$ contains a complete instance of the current state, i.e., a sampled value for each state variable. The current distribution is then approximated by

$$P(X_T = x|Y_0, Y_1, ..., Y_T) = \frac{1}{N}\sum_{i=1}^{N} \delta(x_T^i = x) \qquad (2)$$

where $\delta(x_T^i = x)$ is 1 if the state represented by $x_T^i$ is the same as $x$, and 0 otherwise. The particle filter starts by generating $N$ particles according to the initial distribution $P(X_0)$. Then, at each step, it first generates the next state $x_{t+1}^i$ for each particle $i$ by sampling from $P(X_{t+1}^i|X_t^i)$. It then weights these samples according to the likelihood they assign to the observations, $P(Y_{t+1}|X_{t+1}^i)$, and resamples $N$ particles from this weighted distribution. The particles will thus tend to stay clustered in the more probable regions of the state space, according to the observations.

Although particle filtering has scored impressive successes in many applications, one significant limitation is of special concern: it tends to perform poorly in high-dimensional state spaces. This problem can be reduced by analytically marginalizing out some of the variables, a technique known as *Rao-Blackwellisation* (Murphy & Russell, 2001). When the state space $X_t$ can be divided into two subspaces $U_t$ and $V_t$ such that $P(V_t|U_t, Y_1, \ldots, Y_t)$ can be efficiently computed analytically, we only need to sample from the smaller space $U_t$, and this requires far fewer particles for the same accuracy. Each particle is now composed of a sample from $P(U_t|Y_1, \ldots, Y_t)$ plus a parametric representation of $P(V_t|U_t, Y_1, \ldots, Y_t)$. For example, if the variables in $V_t$ are discrete and independent of each other given $U_t$, we can store for each variable the vector of parameters of the corresponding multinomial distribution (i.e., the probability of each value).

## 3. Relational Dynamic Bayesian Networks

In this section we show how to represent probabilistic dependencies in a dynamic relational domain by combining DBNs with first-order logic. We start by defining relational and dynamic relational domains in terms of first-order logic and then define *relational dynamic Bayesian networks* (RDBNs) which can be used to model uncertainty in such domains.

A relational domain contains a set of objects with relations between them. The domain is represented by constants, variables, functions, terms and predicates. *Constants* are symbols used to represent objects (e.g., $plate_{23}$ can be one of the plate objects in the factory assembly domain) or the attributes of objects (e.g., $red$ is a constant which can be the color of a plate) in the domain. *Variables* range over the objects, and both the constants and the variables can be typed, in which case the variables take on values only of the corresponding type. Functions $f(x_1, \ldots, x_n)$ take objects as arguments and return an object. Functions are associated with an arity $n$ which fixes the number of arguments that the function may take. A *predicate* $R$ is a symbol used to represent relations between objects in the domain or attributes of objects. An *interpretation* specifies which objects, functions and relations in the domain are represented by which symbols. A *term* is an expression used to represent an object in the relational domain. A string $t$ is a term if (a) $t$ is a constant symbol or (b) $t$ is a variable or (c) $t$ is of the form $f(t_1, \ldots, t_n)$ where $f$ is a function and each of the $t_i$ is a term. Each predicate symbol $R$ is associated with an arity $n$, and an *atomic formula* $R(t_1, \ldots, t_n)$ is a predicate symbol applied to an *n-tuple* of terms (e.g., $Weld(x, y)$ means that objects $x$ and $y$ are welded and $Color(x, Red)$ means that the color of object $x$ is $Red$.). A *ground term* is a term containing no variables. A *ground atomic formula* or ground predicate is an atomic formula all of whose arguments are ground terms. Each ground predicate is associated with a truth value and the state of the domain is given by the truth value assigned to all possible ground predicates.

**Definition 1** *(Relational Domain)*
<u>*Syntax:*</u> *A relational domain is a set of constants, variables, functions, terms, predicates and atomic formulas $R(t_1, \ldots, t_n)$ where each of the argument $t_i$ is a term. The set of all possible ground predicates is the set of all predicates with constants (or functions applied to constants) as arguments.*
<u>*Semantics:*</u> *Each ground predicate in a relational domain can be either true or false. The state of a relational domain is the set of ground predicates that are true.*

The set of all true ground predicates can be represented explicitly as tuples in a relational database, and under closed world assumptions this corresponds to a state of the world.

In an uncertain domain, the truth value of a ground predicate can be uncertain and the value can potentially depend on the values of other ground predicates. These dependencies can be specified using a Bayesian network on the ground predicates. However, the number of such ground predicates is exponential in the size of the domain (number of constants) and hence the explicit construction of such a Bayesian network would be infeasible. We use *Relational Bayesian Networks*[1] (RBNs) to compactly represent the uncertainty in the system. The relational Bayesian network specifies the dependency between the predicates at the first-order level by using first-order expressions which include existential and universal quantifiers, and aggregate functions such as *count*, etc.

**Definition 2** *(Relational Bayesian Network: RBN)*
<u>Syntax</u>: *Given a relational domain , a relational Bayesian network is a graph which, for every first-order predicate $R$, contains:*

- *A node in the graph.*

- *A set of parents $Pa(R) = R_1(t_1^1, \ldots, t_{m_1}^1), \ldots, R_l(t_1^l, \ldots, t_{m_l}^1)$ which are a subset of the predicates in the graph (possibly including $R$ itself). The set of parents are indicated by directed edges in the graph from the parent to the child.*

- *A conditional probability model for $P(R|Pa(R))$ which is a function with range [0,1] defined over all the variables in $Pa(R)$.* [2]

<u>Semantics</u>: *A relational Bayesian network defines a Bayesian network on the ground predicates in the relational domain. For every ground predicate $R(c_1, \ldots, c_m)$ a node is created and its parents are obtained by making the substitutions $x_i/c_i$ in the terms $t_k^j$ which appear in the predicate's parent list. The conditional model for a ground predicate is the function restricted to the particular ground predicate and its parents.*

Thus, a relational Bayesian network gives a joint probability distribution on the state of the relational domain.

**Example of an RBN**
Consider a factory assembly process where plates, brackets, etc, are welded and bolted to form complex objects. The plates and the brackets form the objects in the domain. Their properties such as color, shape, etc. can be represented using predicates $Color$, $Shape$, etc. Predicates $Bolt$ and $Weld$ can be used to represent the relationships between the objects. Many of the relationships and the properties of the objects can be uncertain because of faults in the assembly process. For example, a part may be bolted to the wrong part, and this may be more likely if the wrong part and the intended one have the same color. An RBN can model this by having $Color$ as the parent of the $Bolt$ predicate and the conditional probability model can be used to represent the exact dependency.

To avoid cycles appearing in the network obtained after expansion we need to restrict the set of parents of a predicate. To achieve this, we assume an ordering $\prec$ on the predicates and the constants. The ordering forms part of the description of a relational Bayesian network, and is composed of two parts:

- A complete ordering on the predicates in the relational domain.

- A complete ordering on the constants of each type.

---

1. Our RBNs are related to but different from the relational Bayesian networks of Jaeger (1997); see Section 8.
2. Strictly speaking the function is defined over the ground predicates obtained after instantiation. However, for simplicity we have defined it over the first-order predicates.

The ordering $\prec$ between the ground predicates is now given by the following rules:

- $R(x_1, \ldots, x_n) \prec R'(x'_1, \ldots, x'_m)$ if $R \prec R'$.

- $R(x_1, \ldots, x_n) \prec R(x'_1, \ldots, x'_n)$ if there exists an $i$ such that $x_i \prec x'_i$ and $x_j = x'_j$ for all $j < i$ where $x_k$ and $x'_k$ are constants for all $k$.

We now restrict the set of parents of a predicate in a relational Bayesian network as follows:

- The parent set $Pa(R)$ of a predicate $R$ can contain a predicate $R'$ only if either $R' \prec R$ or $R' = R$.

- If $Pa(R)$ contains $R$ then during the expansion $R(x_1, \ldots, x_n)$ has a parent $R(x'_1, \ldots, x'_n)$ only if $R(x'_1, \ldots, x'_n) \prec R(x_1, \ldots, x_n)$.

This ordering implies that in the expanded Bayesian network each ground predicate can only have higher order ground predicates (w.r.t. $\prec$) as parents and hence there cannot be a cycle.

The conditional model can be any first-order conditional model and can be chosen depending on the domain, the model's applicability and ease of use. In this paper, we will be using first-order probability trees (FOPTs) as our conditional model. They can be viewed as a combination of first-order trees (Blockeel & De Raedt, 1998) and probability estimation trees (Provost & Domingos, 2003).

Before defining FOPTs we need to define first-order formulas.

**Definition 3** *(First-order Formula)*
<u>Syntax</u>: *A first-order formula $F$ is of one of the following forms:*

- *an atomic formula $R(t_1, \ldots, t_n)$ where $R$ is a predicate of arity $n$ and each $t_i$ is a term.*

- *$\neg F'$ or $(F' \wedge F'')$ or $(F' \vee F'')$ where $F'$ and $F''$ are first-order formulas.*

- *$\exists x F'$ or $\forall x F'$ where $x$ is a variable and $F$ is a first-order formula.*

- *$\#(= n)x F'$ or $\#(< n)x F'$ or $\#(> n)x F'$ where $x$ is a variable, $F'$ is a first-order formula and $n$ is an integer.*

<u>Semantics</u>: *The semantics of first-order formulas is the same as in standard first-order logic. Additionally, formulas of the form $\#(= n)x F'$ represent a generalized form of quantification. For example, $\#(\geq n)x F'$ is equivalent to the formula $\exists x_1 \cdots x_n F'(x_1) \wedge F'(x_2) \cdots \wedge F'(x_n) \wedge x_1 \neq x_2 \cdots \neq x_n$ and represents the* count *aggregation. Other aggregators such as* max, min, *etc., can also be defined in a similar way.*

**Definition 4** *(First-order Probability Tree: FOPT)*
<u>Syntax</u>: *Given a predicate $R$, and its parents $R_1, \cdots, R_n$, a first-order probability tree (FOPT) is a tree where*

- *Each interior node $n$ contains a first-order formula $F_n$ on the parent predicates.*

- *The child of a node $n$ corresponds to either the true or false outcome of the first-order formula $F_n$.*

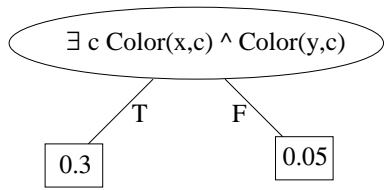

Figure 1: A first-order probability tree for the *Bolted-To(x,y)* predicate.

- *The leaves contain a function with range [0,1] and domain the cross product of all ground parent predicates of R.*

<u>*Semantics*</u>: *An FOPT defines a conditional model for a ground predicate given its parents. The probability is obtained by starting at the root node, evaluating first-order expressions and following the relevant path in the tree to the leaf which encodes the probability in the form of a function.*

An FOPT can contain free variables and quantifiers/aggregators over them. Moreover, the quantification of a variable is preserved throughout the descendants, i.e., if a variable $x$ is substituted by a constant $c$ at a node $n$, then $x$ takes $c$ as its value over all the descendants of $n$. To avoid cycles in the network, quantified variables in an FOPT range only over values that precede the child node's values in the $\prec$ ordering. The function at the leaf gives the probability of the ground predicate being true.

Just like a Bayesian network is completely specified by providing a CPT for each variable, an RBN can be completely specified by having an FOPT for each first-order predicate.

**Example of a first-order probability tree**

Continuing the example of the RBN, the $Bolt$ predicate is dependent on the $Color$ predicate. Figure 1 shows the FOPT for the $Bolt$ predicate. The root node checks the color of the two parts $x$ and $y$. If they have the same color then the probability is 0.3. If they do not have the same color then the probability is 0.05.

We will now consider dynamic relational domains, where the state of the domain changes at every time step. In a dynamic relational domain a ground predicate can be true or false depending on the time step $t$. Therefore we add a time argument to each predicate: $R(x_1, \ldots, x_n, t)$ where $t$ is a non-negative integer variable and indicates the time step.

**Definition 5** *(Dynamic Relational Domain)*
<u>*Syntax*</u>: *A dynamic relational domain is a set of constants, variables, functions, terms, predicates and atomic formulas $R(t_1, \ldots, t_n, t)$ where each of the argument $t_i$ is a term and $t$ is the time step. The set of all possible ground predicates at time $t$ is obtained by replacing the variables in the arguments with constants and replacing the functions with their resulting constants.*
<u>*Semantics*</u>: *The state of a dynamic relational domain at time $t$ is the set of ground predicates that are true at time $t$.*

As before, the dynamic relational domain can contain uncertainty, and specifying the dependencies using a dynamic Bayesian network on the ground predicates is infeasible. We specify the dependencies using a *relational dynamic Bayesian network*. As is the case with dynamic Bayesian networks, we make the assumption that the dependencies are first-order Markov i.e., the predicates at time $t$ can only depend on the predicates at time $t$ or $t - 1$. We also need to add the fact that a grounding at time $t - 1$ precedes a grounding at time $t$:

- $R(x_1, \ldots, x_n, t) \prec R'(x'_1, \ldots, x'_m, t')$ if $t < t'$.

This takes precedence over the ordering between the predicates.

**Definition 6** *(Two-time-slice relational dynamic Bayesian network: 2-TRDBN)*
*A 2-TRDBN is a graph which given the state of the domain at time $t$ gives a distribution on the state of the domain at time $t + 1$. A 2-TRDBN is defined as follows. For each predicate $R$ at time $t$ (i.e, predicate $R$ restricted to groundings at time $t$), we have:*

- *A set of parents $Pa(R) = \{Pa_1, \ldots, Pa_l\}$, where each $Pa_i$ is a predicate at time $t - 1$ or $t$. If $Pa_i$ is at time $t$ then either $Pa_i \prec R$ or $Pa_i = R$.*

- *A conditional probability model for $P(R|Pa(R))$, which is a FOPT on the parent predicates. If $Pa_i = R$, its groundings are restricted to those which precede the given grounding of $R$.*

**Definition 7** *(Relational Dynamic Bayesian Network: RDBN)*
<u>*Syntax:*</u> *A relational dynamic Bayesian network is a pair of networks $(M_0, M_\rightarrow)$, where $M_0$ is an RBN with all $t = 0$ and $M_\rightarrow$ is a 2-TRDBN.*
<u>*Semantics:*</u> *$M_0$ represents the probability distribution over the state of the relational domain at time 0. $M_\rightarrow$ represents the transition probability distribution, i.e., it gives the probability distribution on the state of the domain at time $t + 1$ given the state of the domain at time $t$.*

An RDBN gives rise to a dynamic Bayesian network in the same way that a relational Bayesian network gives a Bayesian network. At time $t$ a node is created for every ground predicate and edges added between the predicate and its parents. (If $t > 0$ then the parents are obtained from $M_\rightarrow$, otherwise from $M_0$). The conditional model at each node is given by the conditional model restricted to the particular grounding of the predicate. We now discuss an example of an RDBN using FOPTs.

**Example of an RDBN**
Consider the factory assembly domain as before where plates, brackets, etc. are welded and bolted to form complex objects. The plates and the brackets have attributes such as shape, size and color. Additionally, a plate can be bolted to a bracket, which is indicated by the *Bolted-To* relation between them. The parts now are assembled over time by performing actions such as painting and bolting which change the attributes of the objects and the relationships between the objects. The actions can be fault-prone which leads to uncertainty and probabilistic dependencies between different attributes. For example, the presence of a bolt between a plate and a bracket might depend on the similarity of their colors, shapes, and other attributes. This can happen because a bolt action can incorrectly bolt objects which are similar to the objects supposed to be bolted. Figure 2 shows the RDBN at time slices $t-1$, $t$ and $t+1$. The nodes in the graph represent the predicates and the edges show the dependencies between them. The predicate *Bolted-To(x,y,t)* represents a bolt between a bracket $x$ and a plate $y$ at time $t$. The predicates $Color(y, c, t)$ and $Shape(y, s, t)$ represent the color and shape of a bracket $y$ with values $c$ and $s$ respectively. The predicate $Bolt(x, y, t)$ represents the bolt action performed between the objects $x$ and $y$ at time $t$. Without loss of generality, we assume that exactly one action is performed per time step. The graph shows that the *Bolted-To* predicate at time $t$ depends on the action performed and the predicates *Bolted-To*, $Color$ and $Shape$ at time $t - 1$. Figure 3 shows the FOPT for the *Bolted-To* attribute. The leaves in the FOPT (represented by the box) contain the probabilities for the predicate being true and the intermediate nodes contain first-order expressions testing the various conditions. The left and the right branches correspond to the expressions being true and false respectively. The FOPT represents the fact that if the bolt between objects $x$ and $y$ existed at time $t - 1$, then it exists at time $t$. Otherwise, if a bolt action was performed on objects $x$ and $y$, then the probability of $x$ and $y$ getting bolted is 0.9. On the other hand, if the bolt action was performed on objects $x$ and $z$, then the probability of $x$ and $y$ getting bolted depends upon the similarity between $y$ and $z$. In this example, two objects are similar if their colors are the same. If $y$ and $z$ are similar, then the probability that $x$ and $y$ get bolted is inversely proportional to the number of similar objects to $z$. We model this by using the $count$ aggregator in the function at the leaf. The expression $count(w|Bracket(w) \wedge Color(w, c, t - 1))$ gives the number of brackets that have the color $c$ (the same as that of $z$) at time $t - 1$. In this example, we allow multiple objects to get bolted due to a single action. However, we might also wish to model mutual exclusion between the ground predicates. This is achieved by creating a special predicate *Mutex(t)* which depends on the ground predicates that are involved in the action performed at the previous time step. Figure 4 shows an FOPT for the *Mutex(t)* predicate. Predicates *Weld(x,y,t)* and *Welded-To(x,y,t)* refer to the weld action and relation respectively. The FOPT represents that if a bolt or weld action was performed, then *Mutex(t)* is true only if at most one additional weld or bolt ground predicate is true at time $t$. During inference *Mutex(t)* is set to true with probability 1 which forces the mutex relation between the ground predicates.

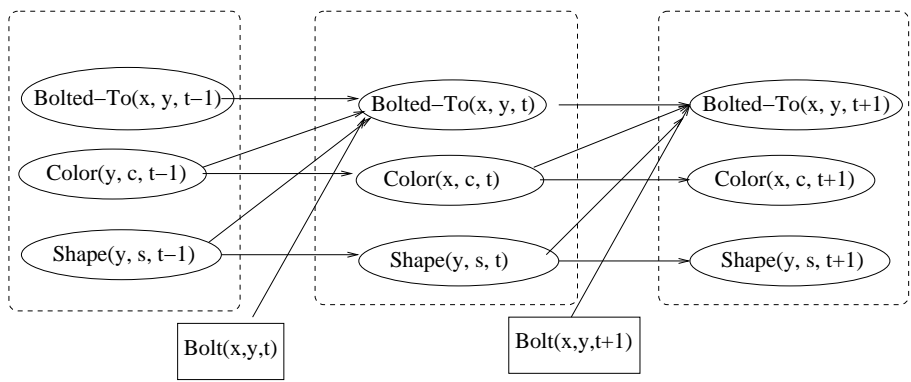

Figure 2: An RDBN representing the assembly domain.

Looking at Figure 3, one might conclude that FOPTs form a tedious representation of the assembly domain. However, this is due to the complex nature of the assembly process and a DPRM modeling the assembly domain would also be quite complex.

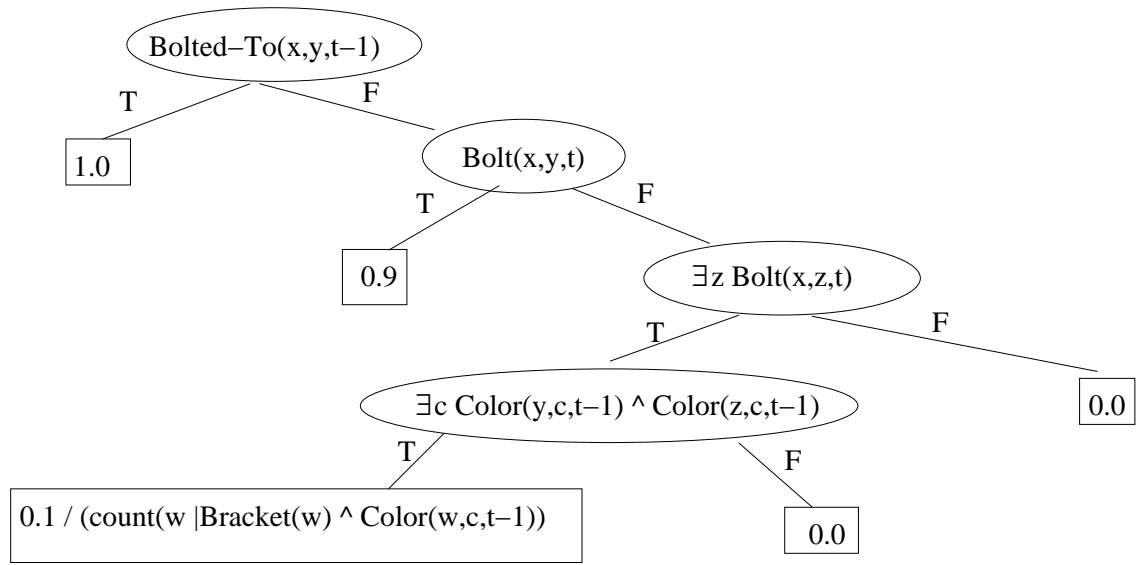

Figure 3: A first-order probability tree for the *Bolted-To(x,y,t)* predicate.

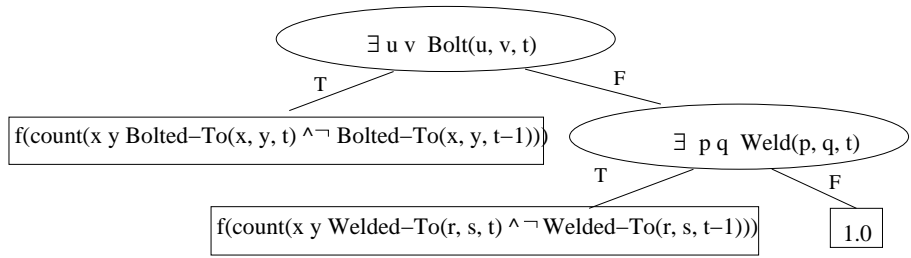

Figure 4: Representing mutual exclusion: an FOPT for *Mutex(t)* predicate; f(a) is 0 if a > 1, else 1.

## 4. Rao-Blackwellized Particle Filtering in RDBNs

This paper addresses the task of state monitoring of a complex relational stochastic process. In the previous section we saw how an RDBN can be used for modeling purposes. In the next few sections we will see how to do efficient inference in RDBNs. As described before, expanding an RDBN gives rise to a DBN. In principle, we can perform inference on this DBN using particle filtering. However, the filter is likely to perform poorly, because for non-trivial RDBNs the state space of the expanded DBN will be extremely large. The DBN will contain a variable for every ground predicate at each time slice and the number of ground predicates is on the order of the size of the domain, which can be in the tens of thousands or more, raised to the arity. We overcome this by adapting Rao-Blackwellisation to the relational setting. We will describe all of our algorithms for predicates with two arguments (excluding the time argument). There generalization to predicates of higher arity is straightforward.

### 4.1 Simple and Complex Predicates

We classify the predicates into two categories, *complex* and *simple*, based on the number, size and types of the arguments. A predicate is termed *complex* if the domain size of the two arguments is large. It is termed *simple* otherwise. Although we do not give a precise definition of large, the intuition becomes clear if we look at the predicates $Color(x, c, t)$ and *Bolted-To(x, y, t)*. The domain size of $c$ in $Color(x, c, t)$ is presumably small, whereas the domain sizes of $x$ and $y$, which represent plates and brackets, could be very large. Hence the particle filter will perform poorly on complex predicates. We now make the following assumptions about the predicates (in later sections we remove them, developing algorithms which can be applied in broader settings).

A1: Uncertain complex predicates[3] do not appear in the RDBN as parents of other predicates.

A2: For any object $o$, there is at most one other object $o'$ such that the ground predicate $R(o, o', t)$ is true. Similarly, there exists exactly one other object $o'$ such that $R(o', o, t)$ is true.

Assumption A1 will, for example, preclude a model where the color of a plate could depend on the properties of brackets bolted to the plate, if the bolt relationship was uncertain. Assumption A2 enforces that a plate can be bolted to at most one bracket (unless we have different predicates for different bolt points). Although these assumptions are restrictive, they still allow us to model many complex domains, and they yield the following very desirable property.

**Proposition 1** *Assumptions A1 and A2 together imply that, given the simple predicates and known complex predicates at times $t$ and $t - 1$, the joint distribution of the unobserved complex predicates at time $t$ is a product of multinomials, one for each predicate.*

Assumption 1 implies that all parents of an unobserved complex predicate are simple or known. Assumption 2 enforces mutual exclusion between the objects participating in a relation with a particular object. Therefore, given a complex first-order predicate and an object, the probabilities of the corresponding ground predicates being true can be seen as a multinomial, with a single trial, over the objects participating in the relation (i.e., the first-order predicate) with the given object. Additionally, the simple predicates are independent of the unknown complex predicates and the complex predicates are independent of each other. Proposition 1 follows.

---

3. These are the predicates whose corresponding ground predicates have uncertain truth values.

Moreover, by assumption A1, unobserved simple predicates can be sampled without regard to unobserved complex ones. Thus, given these assumptions, Rao-Blackwellisation can be applied to speed inference. Recall from Section 2.2 that a Rao-Blackwellised particle is composed of sampled values for all ground simple predicates, plus probability parameters for each complex predicate. The element $R(o_i, o'_j, t)$ stores the probability that the relation holds between objects $o_i$ and $o'_j$ at time $t$ conditioned on the values of the simple predicates in the particle. Rao-Blackwellising the complex predicates can vastly reduce the size of the state space which particle filtering needs to sample. For example, in the FOPT shown in Figure 3, the *Bolted-To* predicate is a complex predicate. Since it only depends on the $Color$ and $Shape$ predicates and the action performed, it satisfies Assumption 1. If, additionally, we do not allow one part to be bolted to more than one part, we can Rao-Blackwellize the *Bolted-To* predicate (and sample the $Color$ and $Shape$ predicates), which can save much space and time.

### 4.2 Memory Efficiency

Even with Rao-Blackwellization, if the relational domain contains a large number of objects and relations, storing and updating all the requisite probabilities can still be quite expensive. This can be ameliorated if context-specific independencies exist, i.e., if a complex predicate is independent of some simple predicates given assignments of values to others (Boutilier, Friedman, Goldszmidt, & Koller, 1996). More precisely, we can group the pairs of objects $(o, o')$ (that can give rise to the ground predicate $R(o, o', t)$) into disjoint sets called *abstractions*: $\mathcal{A}_{\mathcal{R}1}, \mathcal{A}_{\mathcal{R}2}, \cdots, \mathcal{A}_{\mathcal{R}m}$ such that two pairs of objects $(o_i, o'_j)$ and $(o_k, o'_l)$ belong to the same abstraction if $Pr(o_i, o'_j, t) = Pr(o_k, o'_l, t)$. If these relational abstractions can be efficiently specified by first-order logical formulas $\phi$ over the simple predicates, then instead of maintaining probabilities for each pair of objects, we can keep probabilities for each *abstraction*. This can greatly reduce the space required. Similarly, the time required to update the abstractions and the probabilities will be reduced. In Section 7, we run our inference algorithms on the assembly domain. In particular, we model situations where assumptions A1 and A2 are not violated and use the Rao-Blackwellized particle filter for state monitoring (refer Section 7.2). We conclude that it greatly outperforms the standard particle filter (whose number of particles is scaled so that the memory and time requirements are evenly matched). We also observe that using abstractions reduce RBPF's time and memory by a factor of 30 to 70.

### 4.3 Non-functional Relationships

In the remainder of this section, we focus on removing assumption A2, i.e., each object can have relationships with multiple objects via the same relation. However, it is hard to relax this assumption in a way that supports efficient Rao-Blackwellization – if we allow predicates $R(o, o_1, t)$ and $R(o, o_2, t)$ to be simultaneously true we must maintain a joint distribution over these set of ground predicates. As a step towards allowing an arbitrary number of ground relations, first suppose that one is able to bound the number of ground relations per object. Thus, we modify assumption A2 as follows :

A2': For any object $o$, there are at most $\kappa$ objects $o_1, \cdots, o_\kappa$ such that all of $R(o, o_1, t), \ldots, R(o, o_\kappa, t)$ are true (and similarly for the second argument).

In this case, one can maintain a distribution over *sets* of pairs of objects. For example, if the size of the domain of the second argument is $n$, then for every object $o$, the relation $R(o, y, t)$ is true for some $i$ choices of the objects corresponding to the variable $y$, where $i \leq \kappa$. One can maintain a distribution over the various $\binom{n}{i}$ possible combinations for each $i \leq \kappa$. This approach is practical for small $\kappa$, but as $\kappa$ increases the number of sets grows exponentially. In order to reduce the space complexity we group sets having the same probability into an equivalence class where membership is defined with a first-order formula. The formulas and the probabilities may change at time step in accordance with the RDBN and the current state.

Using this abstraction scheme, our experiments show that Rao-Blackwellization can be performed efficiently for $\kappa < 10$ (see Section 7.2).

## 5. Smoothing in RDBNs

Relaxing Assumptions A1 and A2 is difficult because the domains of complex predicates are typically very large, and when complex predicates are parents of other predicates one cannot efficiently compute an analytical solution, rendering Rao-Blackwellization infeasible. If instead the complex predicates are sampled, then an extremely large number of particles is required to maintain accuracy.

Although there is no perfect solution to this problem, we can make use of the fact that similar objects in a relational domain tend to behave similarly, and this similarity extends to the types of relationships in which they participate.

### 5.1 Simple Smoothing

With simple smoothing,[4] we perform standard particle filtering while state monitoring and we will smooth the particles to answer the queries. In this paper, we only describe how to smooth the particles to answer queries about complex predicates. For simple predicates we use the standard particle filter (although our methods are easily extended to them). The simple smoothing approach computes a weighted sum of three components $P_s$, $P_u$ and $P_m$ which we now describe.

The first component $P_s(R(o, o', t))$[5] is the value estimated by a standard particle filter. Here, we sample *all* the ground predicates (both simple and complex). To estimate the probability that $R(o, o', t)$ is true at time $t$, one can count the number of particles in which this relationship holds and divide by the number of particles.

We have already seen that $P_s$ by itself will be inaccurate for any reasonable number of particles due to the curse of dimensionality. To reduce the number of particles needed, we smooth the probability estimate toward two other distributions: $P_u$, the uniform distribution, and $P_m$, the distribution conditioned on the Markov blanket ($MB$).

In the uniform distribution, to compute the probability that $R(o, o', t)$ is true, we ignore the differences between objects, and simply count the fraction of ground predicates for which the relation is true. Thus, the probability that $R(o, o', t)$ is true will be higher if many objects have the relationship between them. The probability is computed as:[6] $P_u(R(o, o', t)) = \frac{1}{N} \sum_{i=1}^{N} \frac{\sum_x \sum_y \delta_i(R(x,y,t),1)}{n_x n_y}$, where $N$ is the number of particles, $\delta_i(R(x, y, t), 1)$ is 1 if $R(x, y, t) = true$ in the $i^{th}$ particle and

---

Bolted–To

| (P1,B1) | (P1,B2) | (P2,B1) | (P2, B2) | (P3,B1) | (P3,B2) |
|---|---|---|---|---|---|
| 0 | 1 | 0 | 0 | 1 | 1 |

| (P1,B1) | (P1,B2) | (P2,B1) | (P2, B2) | (P3,B1) | (P3,B2) |
|---|---|---|---|---|---|
| 0 | 0 | 0 | 1 | 0 | 1 |

Particles

| | | | | | |
|---|---|---|---|---|---|
| 0 | 0 | 1 | 0 | 0 | 0 |
| 0 | 0 | 1 | 0 | 1 | 1 |

Particle Filtering: P(Bolted–To(P1,B1)) = 0

Simple Smoothing: P(Bolted–To(P1,B1)) = 0.8 * 0 + 0.2 * (3/6 + 2/6 + 1/6 + 3/6)/4 = 0.075

Figure 5: An example of simple smoothing.

0 otherwise, and $n_x$ and $n_y$ represent the domain sizes of the first and the second argument of the predicate.

Finally, for each particle, the distribution conditioned on the Markov blanket is obtained by directly computing the probability of the ground predicate given the variable's Markov blanket, which may contain attributes at time $t$ or $t-1$. We then average this estimate over all particles: $P_m(R(o, o', t)) = \frac{1}{N} \sum_{i=1}^{N} P_i(R(o, o', t) \mid MB(R(o, o', t)))$. where $P_i$ represents the probability that $R(o, o', t)$ is true given its Markov blanket $MB$ according to the $i^{th}$ particle. We compute the final probability by smoothing among these three estimates:

$$P(R(o, o', t)) = \alpha_s P_s(R(o, o', t)) + \alpha_u P_u(R(o, o', t)) + \alpha_m P_m(R(o, o', t)) \qquad (3)$$

where $\alpha_s + \alpha_u + \alpha_m = 1$.

We term this approach *simple smoothing*. Figure 5 shows an example of simple smoothing. The weights used are $\alpha_s = 0.8$ and $\alpha_u = 0.2$. For simplicity, we ignore the prediction made by conditioning on the Markov Blanket. There are four particles, each containing sampled values for the *Bolted-To* predicate at some time $t$. The particle filter predicts that *P(Bolted-To(P1,B1,t))* = 0, while simple smoothing predicts the probability to be 0.075. This example highlights the problem with standard particle filtering. There might be a non-zero probability of some ground predicate being true, but due to the large size of the domain particle filter may not have samples which report this.[7] Our experiments show that simple smoothing performs considerably better than standard particle filtering (see Section 7.3), but it overgeneralizes by smoothing over the entire set of objects. We now present a more refined approach based on smoothing over a lattice of abstractions.

---

7. In the experiments, $\alpha_s = 0.8$ and $\alpha_u = \alpha_m = 0.1$. The weights can also be set by adapting the procedure described in the next section.

## 5.2 Abstraction-Based Smoothing

As before, we are interested in computing the marginal probability of $R(o, o', t)$. Instead of using a uniform distribution to smooth the estimates, we can obtain better estimates by only considering the relationship between pairs of objects $o_i, o'_j$ such that $o_i$ is similar to $o$ and $o'_j$ is similar to $o'$. We call a grouping of a set of pairs of objects which are related in some way an *abstraction*. For example, pairs of large plates and brackets which are bolted together form an abstraction. An abstraction will be more general if it allows many objects to be termed as similar. The probability estimates for more general abstractions will be based on more instances, and thus have lower variance, but will ignore more detail, and thus have higher bias. The trade-off is made by using a weighted combination of estimates from a lattice of abstractions. In the next few sections we use the following representation. Each complex predicate $R$ is represented by a set $X_R$ of Boolean indicator variables $X_{l,j}$ where $X_{l,j}$ is 1 if $R(o_l, o'_j, t) = 1$ and 0 otherwise. An abstraction can be thought of as a set of pairs specified by a subset of the indicator variables.

### 5.2.1 LATTICE OF RELATIONAL ABSTRACTIONS

Given a set $S$, a lattice is a set of nodes where each node represents a subset of $S$. In the relational domain we will be building an abstraction lattice over each complex predicate. As described above, we define a *relational abstraction* of $R$ to be a subset of the indicator variables, where the subset contains $X_{l,j}$ if $o_l$ and $o'_j$ satisfy some first-order formula. For example, if the formula is a simple conjunctive expression we have the following. Consider the first-order formula $\phi = A_1(x, u_1, t) \wedge \cdots \wedge A_m(x, u_m, t) \wedge B_1(y, v_1, t) \wedge \cdots \wedge B_m(y, v_n, t)$ where $A_i$ and $B_k$ are simple predicates and $u_i$ and $v_k$ are constants.

**Definition 8** *(Relational Abstraction)*
*The relational abstraction of $R$ specified by $\phi$ is the set $\mathcal{A}_\mathcal{R} \subseteq X_R$ defined as:*

$$\mathcal{A}_\mathcal{R} = \{X_{l,j} \in X_R | A_1(o_l, u_1, t) \wedge \cdots \wedge A_m(o_l, u_m, t) \wedge B_1(o'_j, v_1, t) \wedge \cdots \wedge B_n(o_j, 'v_n, t)\}$$

As an example of an abstraction, consider the assembly domain and the predicate *Bolted-To*. An abstraction could be $\phi = Color(x, red, t) \wedge Size(y, large, t)$ which represents the *Bolted-To* relation between all *plates* which are *red* and all *brackets* which are *large*. These relational abstractions will form a lattice and our goal is to use the relational abstraction lattice along with smoothing to improve particle filtering.

### 5.2.2 SMOOTHING WITH AN ABSTRACTION LATTICE

Given a relational attribute $R$, we have to estimate the probability of $R(o, o', t)$ being true, which we shall refer to as $P(X_{a,b} = 1)$, where $o$ and $o'$ have the indices $a$ and $b$ in $X_R$. We smooth the particle filter estimates over the relevant abstractions of $R$. Given the RDBN and the ground predicate $R(o, o', t)$, we first consider the set of parents $(R_1, \cdots, R_n)$ of the predicate. Given the $i^{th}$ particle, for each of the parents, we set a value which is either equal to the parent's value in the particle or $*$ *(don't care)*. This defines a relevant abstraction. More formally, an abstraction is relevant to $R(o, o', t)$ if it is a conjunctive expression involving a subset $R$'s parents, and their values in the expression are the values in some particle $i$ (or, more generally, if $(o', o, t)$ satisfies some first-order expression; conjunction is a special case). The intuition behind using such abstractions is that if the set of parents of some variables are the same and the parents have exactly the same values, then the probability of a variable being true can be obtained by looking at the distribution of the variables in the particles.

Thus for each subset of parent attributes and their corresponding values we have an abstraction. If we consider all possible subsets of the parents, we obtain a lattice of abstractions.

For example, if the presence of a bolt between a plate and a bracket depends on the size of the bracket and the size and color of the plate but not on other attributes, then the abstraction lattice is defined over the size of the bracket and the size and color of the plate. If the color of $o$ is red and the size of $o'$ is large then the relational abstraction which will be used in smoothing to find $P(R(o, o', t))$ will have a subset of the *color* and *size* predicates and the values specified as above. Figure 6 shows an example of an abstraction lattice. The first-order formula describing the abstractions can also contain complex predicates and quantifiers/aggregators over them.

The number of abstractions is exponential in the number of parents. If the number becomes too large, we use an approach based on rule induction to select the most informative abstractions (see Algorithm 1). For each abstraction, we first define a score which is the K-L divergence (Cover & Thomas, 2001) between the distribution of $R$ predicted by the abstraction and the empirical distribution. The empirical distribution is obtained by taking all the ground instances $R(o, o', t)$ across all the particles. For simplicity, we assume that instances within a particle are independent and identically distributed (although this may not be the case). These instances can then be considered as independent samples of the true distribution. The distribution $\hat{p}_{\mathcal{A}_\mathcal{R}}$ predicted by an abstraction $\mathcal{A}_\mathcal{R}$ is obtained by averaging across ground instances which belong to the abstraction, i.e.,

$$\hat{p}_{\mathcal{A}_\mathcal{R}}(x) = \frac{\sum_i \sum_{X_{l,j} \in \mathcal{A}_\mathcal{R}} \delta^i(x_{l_j}, x)}{N|\mathcal{A}_\mathcal{R}|}$$

where $x$ is either 0 or 1, $N$ is the number of particles, $|\mathcal{A}_\mathcal{R}|$ is the size of the abstraction, and $\delta^i(x_{l_j}, x) = 1$ if the value $x_{l_j}$ of the indicator variable $X_{l,j}$ is $x$ in the $i^{th}$ particle and 0 otherwise.

We approximate the K-L divergence between the empirical distribution $p$ and $\hat{p}_{\mathcal{A}_\mathcal{R}}$ as described in Section 7.1:

$$score(\mathcal{A}_\mathcal{R}) = \hat{D}_H(p||\hat{p}_{\mathcal{A}_\mathcal{R}}) = -\frac{1}{N|\mathcal{A}_\mathcal{R}|} \sum_i \sum_{X_{l,j} \in \mathcal{A}_\mathcal{R}} \log \hat{p}(x_{l_j})$$

Algorithm 1 shows the procedure to select the most relevant abstractions. We start off with the null abstraction (i.e., the most general abstraction) and greedily add an attribute-value pair to it which maximizes the *score* function. (The attributes correspond to the parent predicates.) We keep on adding attribute-value pairs until either the *score* cannot be improved or the number of attribute-value pairs (also termed the length of the abstraction) exceeds the *maxLen* parameter (to prevent overfitting). We then add the new abstraction to the list of relevant abstractions. To avoid redundancy among the abstractions, we remove ground instances that the new abstraction covers. We then repeat the procedure with the updated set of ground instances. When the number of abstractions exceeds the maximum number, we stop the search. Pruning can also be done by using a holdout set of particles and evaluating the abstractions' scores on them, or any of the other methods used in rule induction (see Clark & Niblett, 1989; Cohen, 1995, etc.).

In our experiments, there were typically only a few parents, so we used all possible abstractions.

---

**Algorithm 1** Abstraction Lattice Smoothing.

---

$Pa(R) \leftarrow (R_1, \cdots, R_n)$
$null\mathcal{A}_{\mathcal{R}} \leftarrow \{\}$
$RelevantAbs \leftarrow \{\}$
$n \leftarrow 0$
**while** $n < maxAbs$ **do**
   $current\mathcal{A}_{\mathcal{R}} \leftarrow null\mathcal{A}_{\mathcal{R}}$
   $minKLD \leftarrow \infty$
   $len \leftarrow 0$
   **while** $len \leq maxlen$ **do**
      **for all** $R_i \in Pa(R) \setminus current\mathcal{A}_{\mathcal{R}}$ **do**
         **for all** $V_j \in Dom(R_i)$ **do**
            $temp\mathcal{A}_{\mathcal{R}} \leftarrow current\mathcal{A}_{\mathcal{R}} \cup (R_i, V_j)$
            **if** $temp\mathcal{A}_{\mathcal{R}} \in RelevantAbs$ **then**
               continue
            **end if**
            $KLD \leftarrow score(temp\mathcal{A}_{\mathcal{R}})$
            **if** $KLD < minKLD$ **then**
               $new\mathcal{A}_{\mathcal{R}} \leftarrow temp\mathcal{A}_{\mathcal{R}}$
               $minKLD \leftarrow KLD$
            **end if**
         **end for**
      **end for**
      **if** $new\mathcal{A}_{\mathcal{R}} = current\mathcal{A}_{\mathcal{R}}$ **then**
         exit
      **else**
         $current\mathcal{A}_{\mathcal{R}} \leftarrow new\mathcal{A}_{\mathcal{R}}$
         $len \leftarrow len + 1$
      **end if**
   **end while**
   $n \leftarrow n + 1$
   Add $current\mathcal{A}_{\mathcal{R}}$ to $RelevantAbs$
   Remove ground instances of $R$ covered by $current\mathcal{A}_{\mathcal{R}}$
**end while**

---

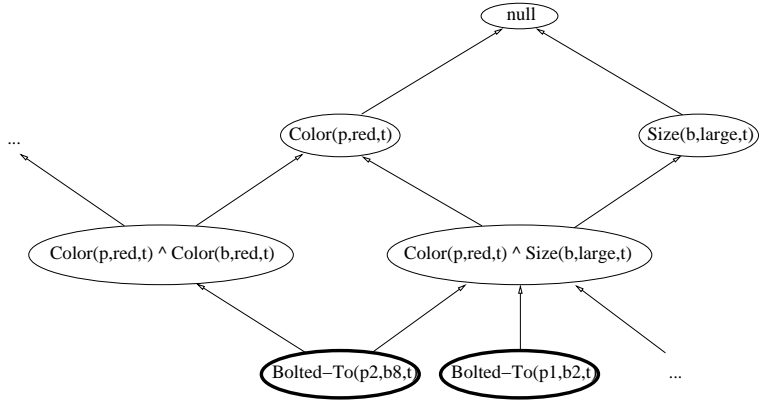

Figure 6: An abstraction lattice over the *Bolted-To* relation between $plate_1$ (red) and $bracket_2$ (large).

The abstractions obtained are then assigned weights by adopting the heuristic *length* formula proposed by Anderson et al (2002) (where it is called the *rank* heuristic). The weight $\omega_{\mathcal{A}_{\mathcal{R}}}$ of the abstraction $\mathcal{A}_{\mathcal{R}}$ is computed as:

$$\omega_{\mathcal{A}_{\mathcal{R}}} \propto |\mathcal{A}_{\mathcal{R}}|^{\text{Length}(\mathcal{A}_{\mathcal{R}})}$$

where $|\mathcal{A}_{\mathcal{R}}|$ is the size of the abstraction, i.e., the number of ground predicates that can belong to $\mathcal{A}_{\mathcal{R}}$, and $\text{Length}(\mathcal{A}_{\mathcal{R}})$ is the length of the abstraction. The intuition behind this formula is to trade off bias and variance by giving more weight to abstractions with many samples, but less weight to abstractions that are overly general. We also tried using the EM algorithm to learn the weights, as described by McCallum et al (1998), by using the various particles as the data points. In our experiments, we found that both work well in practice, but the heuristic length formula is much more efficient.

The probability estimate $P(X_{a,b} = 1)$ is a weighted average of the probability estimates given by the various abstractions:

$$P_i(R(o, o', t)) = P_i(X_{a,b} = 1) = \frac{1}{c} \sum_{\mathcal{A}_{\mathcal{R}} \ni X_{a,b}} \omega_{\mathcal{A}_{\mathcal{R}}} \frac{n^i_{\mathcal{A}_{\mathcal{R}}}}{|\mathcal{A}_{\mathcal{R}}|} \qquad (4)$$

where $P_i$ is the probability as represented by the $i^{th}$ particle, $c = \sum_{\mathcal{A}_{\mathcal{R}}} \omega_{\mathcal{A}_{\mathcal{R}}}$ is a normalizing constant, $\mathcal{A}_{\mathcal{R}}$ is a relational abstraction that has $X_{a,b}$ as one of its elements, $n^i_{\mathcal{A}_{\mathcal{R}}}$ is the number of indicator variables belonging to $\mathcal{A}_{\mathcal{R}}$ which have the value 1 in the $i^{th}$ particle, $|\mathcal{A}_{\mathcal{R}}|$ is the size of the abstraction (i.e., the number of indicator variables in $\mathcal{A}_{\mathcal{R}}$), and $\omega_{\mathcal{A}_{\mathcal{R}}}$ is the weight of the abstraction $\mathcal{A}_{\mathcal{R}}$.

$P_i$ is computed for each particle and the final probability is given by the average of $P_i$ over all particles:

$$P(X_{a,b} = 1) = \frac{\sum_{i=1}^{N} P_i(X_{a,b} = 1)}{N} \qquad (5)$$

where $N$ is the number of particles.

Thus, particle filtering proceeds as usual, except that during inference, at each time step, the probabilities are computed using the above formula. The experiments (Section 7) show that this method greatly outperforms standard particle filtering and simple smoothing.

## 6. Relational Kernel Density Estimation

One way to approximate the joint distribution of the relational variables is to assume independence between all the indicator variables corresponding to each ground predicate. The marginal probabilities computed above can then be used to compute the joint distribution as:

$$P(X = x) = \prod_{R \in X} \left[ \prod_{X_{l,j} \in X_R} P(X_{l,j} = x_{l,j}) \right] \qquad (6)$$

where $X$ represents the joint state variable, $x$ is a particular state, $R$ is a predicate and $X_R$ is the corresponding set of indicator variables $x_{l,j}$ and $P(X_{l,j} = x_{l,j})$ is given by Equation 5. This formula only describes the joint distribution of the complex predicates, but can be easily extended to compute the joint distribution of all the predicates. However, this approach can lead to inaccurate results when the independence assumption does not hold. Moreover, the marginal probability has to be calculated for every indicator variable (i.e., ground predicate) irrespective of whether its value in the state is true or not, and working in such a high-dimensional space makes this inefficient.

In this section we propose a form of kernel density estimation (Duda, Hart, & Stork, 2000) to directly compute the joint probability distribution of the variables efficiently and accurately. A kernel density estimator for a variable $X$ takes $n$ samples (i.e., particles $x^i$) and estimates $X$'s probability distribution as:

$$P(X = x) = \frac{1}{n} \sum_i K(x, x^i)$$

where $K$ is a non-negative kernel function that satisfies $\sum_x K(x, x^i) = 1$, for all $i$. The kernel function $K$ represents a distribution over $X$ based on the sample $x^i$ and is typically a function of the distance between $x^i$ and $x$. For example, if $x$ and $x^i$ are Boolean vectors, then the distance $d(x, x^i)$ can be the Hamming distance between the vectors, and $K(x, x^i) = \frac{1}{d(x,x^i)^2}$. However, in our case this is unlikely to give good results because kernel density estimation usually does not work well in high dimensions. To overcome this problem we first break our kernel function into a product of kernel functions, one for each complex predicate. Thus we have:

$$K(x, x^i) = \prod_R K_R(x, x^i) \qquad (7)$$

However, each of the complex predicates, when viewed as a Boolean vector, can itself be very high-dimensional and sparse, leading to $d(x, x^i)$ being the same for most $(x, x^i)$ pairs, and producing poor results. Fortunately, the sparsity can itself be used to reduce the effective dimension of the kernel function for a relation. Let $n_{X_{l,j}=1}(X_R)$ represent the cardinality of the subset of indicator variables that have the value 1. We divide the kernel function into two factors. The first factor of $K_R$ gives the probability distribution on the number of indicator variables whose value is 1 given the number of indicator variables whose value is 1 in particle $x^i$, i.e., $P(n_{x_{l,j}=1}(x_R) | n_{x_{l,j}^i=1}(x_R^i))$. The particles are erroneous in reproducing the exact relationships in the domain, but they can approximately capture the *number of relationships*. Thus, we model this number using a binomial distribution where the number of trials is $n_{x^i_{l,j}=1}(x^i_R)$, each with a success probability of $p_s$. The parameter $p_s$ is computed from the model. For example, in the assembly domain $p_s$ will depend on the fault probability of the actions. The binomial model is used because of mutual exclusion in the assembly domain, which causes the number of true ground predicates to be approximately equal to the number of actions performed. In other domains, one may not require this factor.

The second factor of $K_R$ is the average probability of the indicator variables that are 1 in the state, given the indicator variables that are 1 in the particle. To estimate this, we can once again use the abstractions, in particular Equation 4.

However, the average of these probabilities will generally not sum to 1 over the various states. Hence, to make $K_R$ a kernel function we must normalize it over all the possible substates $X$ such that $n_{X_{l,j}=1}(X_R) = n_{x_{l,j}=1}(x_R)$. In conclusion our kernel function is

$$K_R(x, x^i) = B(n_{x_{l,j}=1}(x_R), n_{x^i_{l,j}=1}(x^i_R), p_s) \frac{\sum_{x_{a,b} \in S} P_i(x_{a,b}=1)}{d \, n_{x_{l,j}=1}(x_R)} \tag{8}$$

where $B(k, n, p)$ represents the binomial distribution (probability of $k$ successes in $n$ trials with success probability $p$), $d$ is the normalization factor, $S$ is the subset of indicator variables with value 1, $n_{x_{l,j}=1}(x_R)$ is the cardinality of $S$, and $P_i()$ is given by Equation 4.

Computing the normalization factor by summing over the various states as described above can be exponential in the number of ground predicates present in the state and thus infeasible. However, in our case the normalization factor can be computed analytically and is given by Proposition 2.

**Proposition 2** *The normalization factor* $d = \binom{|X_R|-1}{n_{x_{l,j}=1}(x_R)-1} \sum_{\mathcal{A}_R} \frac{\omega_{\mathcal{A}_R} n^i_{\mathcal{A}_R}}{n_{x_{l,j}=1}(x_R)}.$

*Proof*: For convenience we use $n(x_R)$ instead of $n_{x_{l,j}=1}(x_R)$.

$$
\begin{aligned}
d &= \sum_{y_R \in X_R : n(y_R)=n(x_R)} \quad \sum_{y_{l,j} \in y_R : y_{l,j}=1} \frac{\sum_{\mathcal{A}_R : y_{l,j} \in \mathcal{A}_R} \frac{w_{\mathcal{A}_R} n^i_{\mathcal{A}_R}}{|\mathcal{A}_R|}}{n(x_R)} \\
&= \sum_{\mathcal{A}_R} \sum_{y_{l,j} \in \mathcal{A}_R} \sum_{y_R : y_{l,j}=1, n(y_R)=n(x_R)} \frac{w_{\mathcal{A}_R} n^i_{\mathcal{A}_R}}{|\mathcal{A}_R| n(x_R)} \\
&= \sum_{\mathcal{A}_R} \sum_{y_{l,j} \in \mathcal{A}_R} \binom{|X_R|-1}{n(x_R)-1} \frac{w_{\mathcal{A}_R} n^i_{\mathcal{A}_R}}{|\mathcal{A}_R| n(x_R)} \\
&= \sum_{\mathcal{A}_R} |\mathcal{A}_R| \binom{|X_R|-1}{n(x_R)-1} \frac{w_{\mathcal{A}_R} n^i_{\mathcal{A}_R}}{|\mathcal{A}_R| n(x_R)} \\
&= \binom{|X_R|-1}{n(x_R)-1} \sum_{\mathcal{A}_R} \frac{w_{\mathcal{A}_R} n^i_{\mathcal{A}_R}}{n(x_R)}
\end{aligned}
$$

□

Figure 7 shows a hypothetical example consisting of three particles which contain the sampled values for the *Bolted-To* ground predicates. The example also describes the abstraction lattice which in this case is a tree. The probability of the state *(0,0,0,0,0,0,0,1)* as predicted by a standard particle filter is 0.0 because none of the particles contain this state. The joint probability computed using independence assumptions requires calculating the probability for each ground predicate separately and multiplying these probabilities. For example, the abstraction smoothing method gives a probability of *Bolted-To(P2, B1)* being false of $0.8 * 2/3 + 0.2 * 9/12 = 0.7$. The overall probability is given in the figure. This probability can be highly inaccurate as well as expensive to compute. Finally, relational kernel density estimation computes the probability only for the ground predicates that are true (i.e., 1) and averages them out. Using the abstraction smoothing we can see that this probability comes out to be $0.8 * 0 + 0.2 * 3/12 = 0.05$. As described above the kernel is also multiplied by the binomial distribution $B$ which is the probability of the number of true ground predicates in the state given the number of true ground predicates in the particle (and the RDBN model). In this case, if we assume that the fault probability is low, then $p_s$ will be close to 1 and since all the particles have exactly one ground predicate which is true, $B$ will be close to 1. Hence, the kernel method will predict that the state has a probability of 0.05.

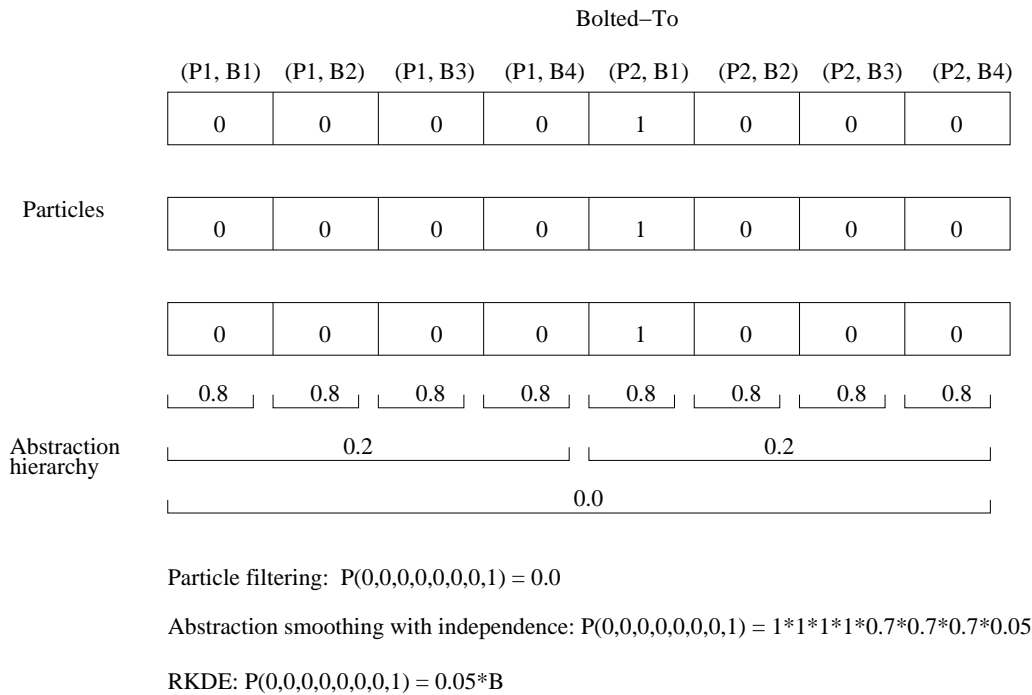

Figure 7: An example of computing joint distributions of complex predicates by particle filtering, abstraction smoothing and RKDE.

## 7. Experiments

In this section we study the application of RDBNs to fault detection in complex assembly plans. We first describe the domain and the experimental procedure we used to study the performance of the various algorithms.

## 7.1 Experimental Setup

We use a modified version of the *Schedule World* domain from the AIPS-2000 Planning Competition (Bacchus, 2001). The problem consists of generating a plan for assembly of objects with operations such as painting, polishing, etc. (see Appendix B for the exact details). Each object has attributes such as surface type, color, hole size, etc. We add two relational operations to the domain: bolting and welding. We assume that actions may be faulty, with fault model described below. In our experiments, we first generate a plan using the FF planner (Hoffmann & Nebel, 2001) assuming that the actions are deterministic (i.e., have no faults). We then monitor the plan's execution explicitly considering possible faults using particle filtering (PF), Rao-Blackwellised particle filtering (RBPF), particle filtering with simple smoothing (SPF), particle filtering with smoothing using an abstraction lattice (ASPF), and particle filtering using relational kernel density estimation (RKDE).

We consider three types of objects: *Plate*, *Bracket* and *Bolt*. *Plate* and *Bracket* have attributes such as weight, shape, color, surface type, hole size and hole type, while *Bolt* has attributes such as size, type and weight. Plates and brackets can be welded to each other or bolted to bolts. The constants in the domain represent objects (e.g., $plate_{23}$) and values of attributes (e.g., $red$). Attributes and relationships represent binary predicates. Actions such as painting, drilling and polishing change the values of the attributes of an object. The action *Bolt* creates a bolt relation between a *Plate* or *Bracket* object to a *Bolt* object. The *Weld* action welds a *Plate* or *Bracket* object to another *Plate* or *Bracket* object. The actions are fault-prone; for example, with a small probability a *Weld* action may have no effect or may weld two incorrect objects based on their similarity to the original objects. This gives rise to uncertainty in the domain and the corresponding dependence model for the various attributes. The fault model has a global parameter, the *fault probability* $p_f$. With probability $1 - p_f$, an action produces the intended effect. With probability $p_f$, one of several possible faults occurs. Faults include a painting operation not being completed, the wrong color being used, the polish of an object being ruined, etc. The probability of these faults depends on the properties of the object being acted on. In addition there are faults such as bolting the wrong objects and welding the wrong objects. The probability of choosing a particular wrong object depends on its similarity to the intended object. Similarity depends on the propositional attributes of the objects involved. Thus the probability of a particular wrong object being chosen is uniform across all objects with the same relevant attribute values.

We allow each object to be "attached" to several other objects. The *Bolt* and *Weld* actions can attach two objects depending on previously attached objects and/or their properties. For example, a plate $A$ may be welded to plate $B$ if both of them are already welded to a common plate. These together violate assumptions A1 and A2 and serve as a good testbed for the various inference algorithms.

The relational process also includes a noisy observation model. When an action is performed on one or more objects, all the ground predicates involving these objects are observed, and no others. With probability $1 - p_o$ the true value of the attribute is observed, and with probability $p_o$ an incorrect value is observed. In our experiments, we set $p_o = p_f$.

A natural measure of the accuracy of an approximate inference procedure is the K-L divergence between the distribution it predicts and the actual one (Cover & Thomas, 2001). However, computing the K-L divergence requires performing exact inference, which for non-trivial RDBNs is infeasible. Thus we estimate the K-L divergence by sampling, as follows. Let $D(p||\hat{p})$ be the K-L divergence between the true distribution $p$ and its approximation $\hat{p}$, and let $\mathcal{X}$ be the domain over which the distribution is defined. Then

$$D(p||\hat{p}) \stackrel{\text{def}}{=} \sum_{x \in \mathcal{X}} p(x) \log \frac{p(x)}{\hat{p}(x)} = \sum_{x \in \mathcal{X}} p(x) \log p(x) - \sum_{x \in \mathcal{X}} p(x) \log \hat{p}(x)$$

The first term is simply the entropy of $X$, $H(X)$, and is a constant independent of the approximation method. Since we are mainly interested in measuring differences in performance between approximation methods, this term can be neglected. The K-L divergence can now be approximated by taking $S$ samples from the true distribution:

$$\hat{D}_H(p||\hat{p}) = -\frac{1}{S} \sum_{i=1}^{S} \log \hat{p}(x_i)$$

where $\hat{p}(x_i)$ is the probability of the $i^{th}$ sample according to the approximation procedure, and the $H$ subscript indicates that the estimate of $D(p||\hat{p})$ is offset by $H(X)$. We thus evaluate the accuracy of particle filtering (PF) and other algorithms on an RDBN by generating $S = 10,000$ sequences of states and observations from the RDBN, passing the observations to the particle filter, inferring the marginal probability of the sampled value of each state variable at each step, plugging these values into the above formula, and averaging over all variables. Notice that $\hat{D}_H(p||\hat{p}) = \infty$ whenever a sampled value is not represented in any particle. The empirical estimates of the K-L divergence we obtain will be optimistic in the sense that the true K-L divergence may be infinity, but the estimated one will still be finite unless one of the values with zero predicted probability is sampled. This does not preclude a meaningful comparison between approximation methods, however, since on average the worse method should produce $\hat{D}_H(p||\hat{p}) = \infty$ earlier in the time sequence. We thus report both the average K-L divergence before it becomes infinity and the time step at which it becomes infinity, if any.

## 7.2 RBPF vs PF

First, we compare the Rao-Blackwellized particle filter (RBPF) with the standard filter (PF) in the case where assumptions A1 and A2 hold. Figure 8 shows the comparison for 1000 objects and varying fault probabilities. The graph shows the K-L divergence at every 100th step. The error bars are the standard deviations. Graphs are interrupted at the first point where the K-L divergence became infinite in any of the runs (once infinite, the K-L divergence never went back to being finite in any of the runs), and that point is labeled with the average time step at which the blow-up occurred. We allocated PF far more particles (200,000) than RBPF (5,000) so that the memory and time requirements are approximately the same for both techniques. As can be seen, for all fault probabilities, PF tends to diverge rapidly, while the K-L divergence of RBPF increases only very slowly. We have also run experiments where the number of objects is varied from 500 to 1500. As can be seen from Figure 9, RBPF outperforms PF in this case as well.

Next we compare RBPF with PF when the number of objects that can be attached to a particular object is greater than one. The maximum number of relationships per object that we consider is 10. From Figure 10, we can conclude that RBPF performs equally well in this case. Although RBPF gives quite accurate predictions, its speed decreases as the number of relationships increases. Figure 11 confirms this and shows the need for faster algorithms. In all the above experiments we use object abstractions (Section 4) which reduce RBPF's time and memory by a factor of 30 to 70, and take on average six times longer and 11 times the memory of PF, per particle. However, note that we run PF with 40 times more particles than RBPF. Thus, RBPF is using less time and memory than PF, while predicting behavior far more accurately.

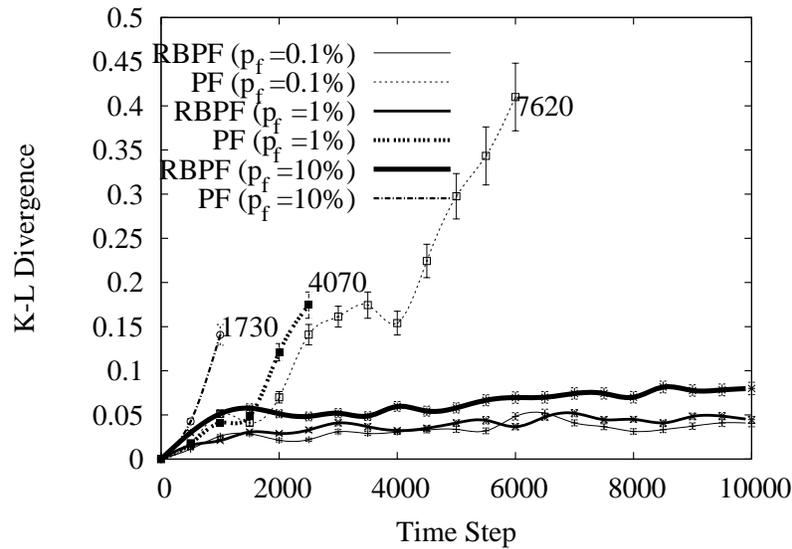

Figure 8: RBPF (with 5,000 particles) has much less error than standard PF (with 200,000 particles) in domains where assumptions A1 and A2 are not violated. This experiment was done with 1000 objects and varying fault probabilities.

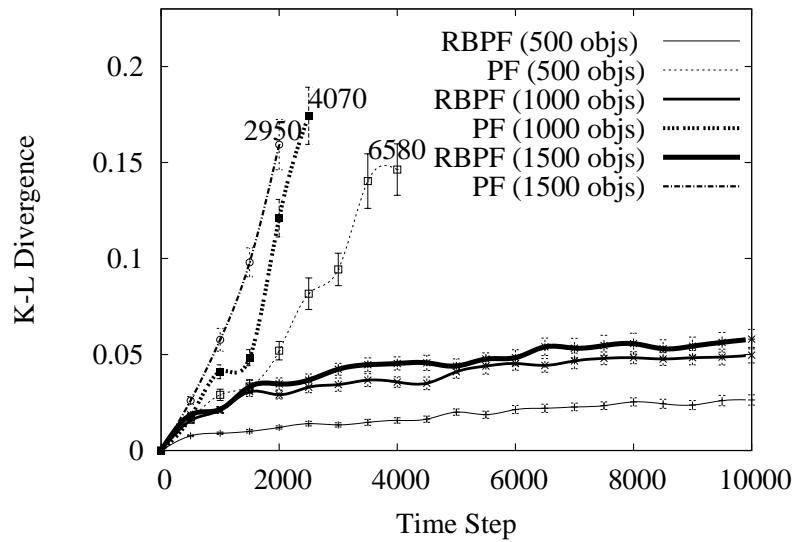

Figure 9: RBPF outperforms PF for varying numbers of objects. The fault probability in this experiment was kept constant at $p_f = 1\%$.

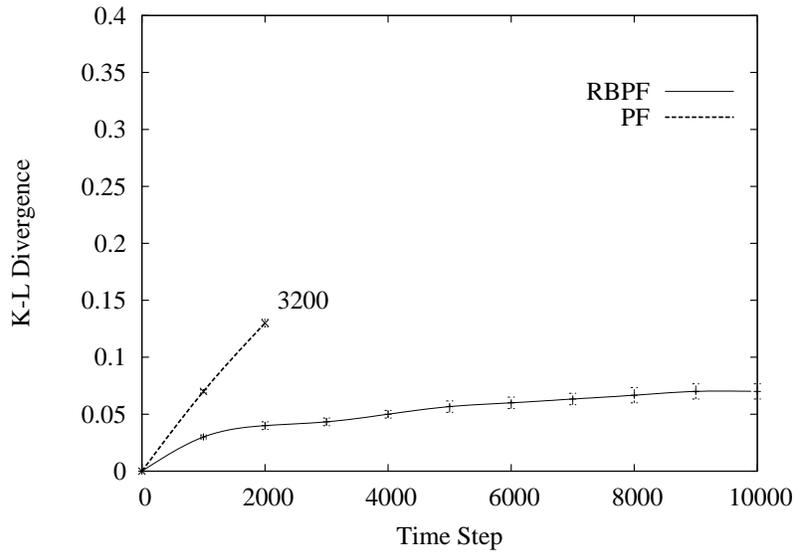

Figure 10: The graph shows the performance of RBPF when the maximum number of relationships per object is increased from 1 to 10. RBPF outperforms the scaled PF for 1000 objects and $p_f = 1\%$.

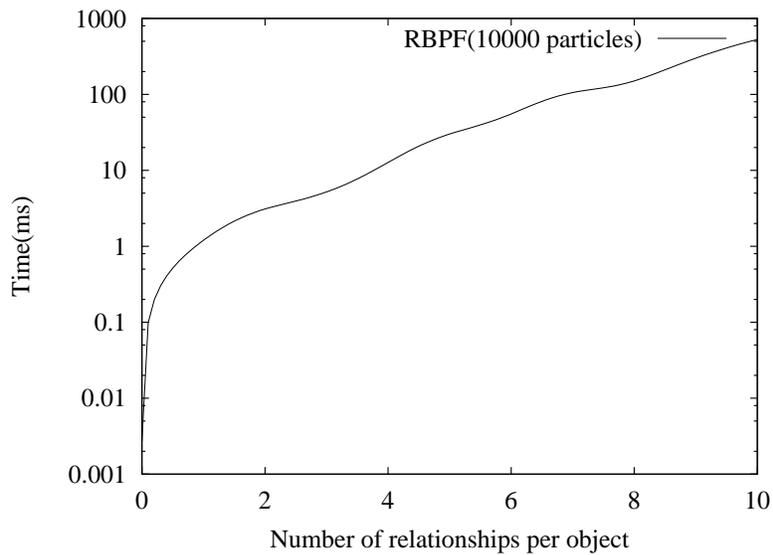

Figure 11: The time taken by RBPF (plotted in log-scale) increases exponentially with the number of relationships per object.

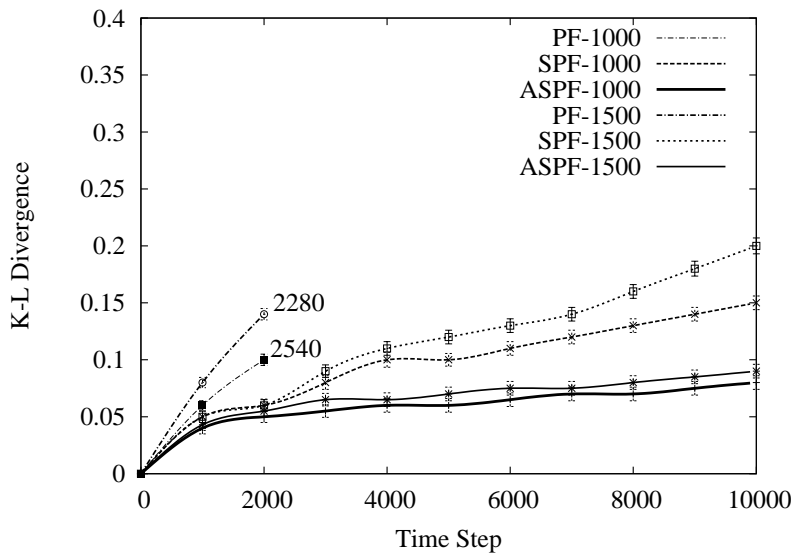

Figure 12: ASPF (20,000 particles) greatly outperforms standard PF (100,000 particles) and SPF (50,000 particles) in predicting the marginal distributions for 1000 and 1500 objects and $p_f = 1\%$.

## 7.3 Computing Marginal Distributions

RBPF works only when Assumption A1 holds (i.e., no predicate depends on uncertain complex predicates) and it becomes slower when Assumption A2 is removed (i.e., relationships are no longer one-to-one). We now study the performance of PF, PF with simple smoothing (SPF) and PF with abstraction-based smoothing (ASPF) which can be used to obtain the marginal distribution when Assumptions A1 and A2 are removed. Figure 12 shows the K-L divergence of the algorithms at every 100th step on an experiment performed with 1000 and 1500 objects and $p_f = 1\%$. Our algorithms use the same amount of memory as PF, but require additional time (on average by a factor of 2 and 5 respectively) to do smoothing. Thus, in our experiments the number of particles used by standard PF, SPF and ASPF are 100,000, 50,000 and 20,000 respectively. One can see that PF tends to diverge very quickly (even with many more particles), while ASPF performs best and its approximation to the marginal distribution is close to the actual distribution. Although the abstraction smoothing algorithm has low error, we observe in the graph that the error increases with time. We attribute this growth to the fact that the effective dimension of the assembly domain increases over time as new (possibly faulty) relations are created, making it increasingly difficult to approximate the distribution with a fixed number of particles. Figure 13 shows the results of experiments for varying fault probabilities. From the two figures we can conclude that the performance of the standard PF degrades with increasing fault probability and with the number of objects, while ASPF remains almost unaffected.

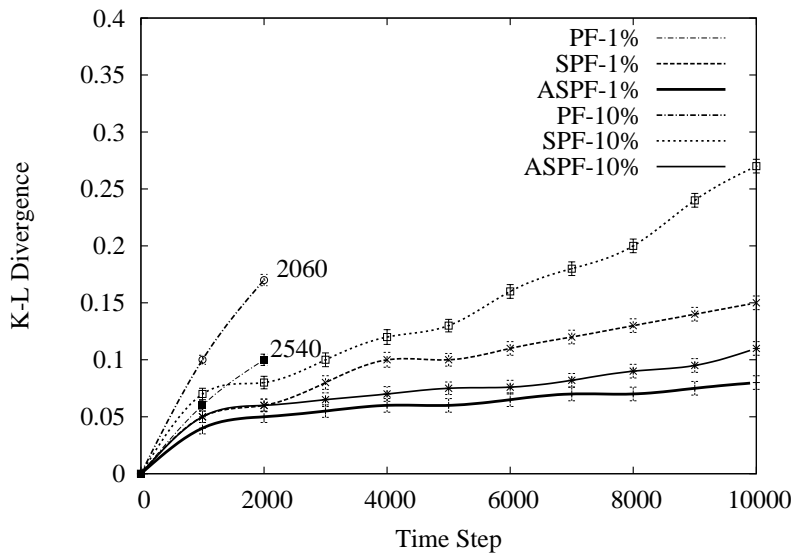

Figure 13: ASPF predicts the marginal distributions most accurately for varying fault probabilities (1%, 10%) and 1000 objects.

Next, we report experiments when Assumption A1 holds and compare ASPS with Rao-Blackwellized particle filtering. The experiment was performed on 1000 objects with fault probability $p_f = 1$ %. Figure 14 shows the mean K-L divergence between the approximate marginal distributions and the true ones. We can see that the difference between the K-L divergence of ASPS and the K-L divergence of RBPF is very small and this difference remains almost constant over time. We conclude that the approximations underlying abstraction smoothing are quite good. Figure 14 shows that the K-L divergence for ASPF is greater than RBPF by at most 0.01, and is on average around 0.005, indicating that our approximations are quite good. We also compare RBPF and ASPF when the number of relationships per object is increased. Figure 15 plots the K-L divergence at the last step when the two algorithms are run on 1000 objects and 1% $p_f$ and varying number of objects per slot. One can see that the difference is always less than .03 and the curve is quite stable.

## 7.4 Computing Joint Distributions

Figures 16 and 17 show the K-L divergence of the full joint distribution of the state (as opposed to just the marginals) for PF, PF with abstraction smoothing (using independence assumptions) and PF using relational kernel density estimation (RKDE) on experiments done with varying number of objects and fault probabilities respectively. One can see that the estimates for the joint probability have greater K-L divergence (which is expected) and RKDE gives the best results. From the experiments we conclude that RKDE can estimate the joint distribution quite accurately.

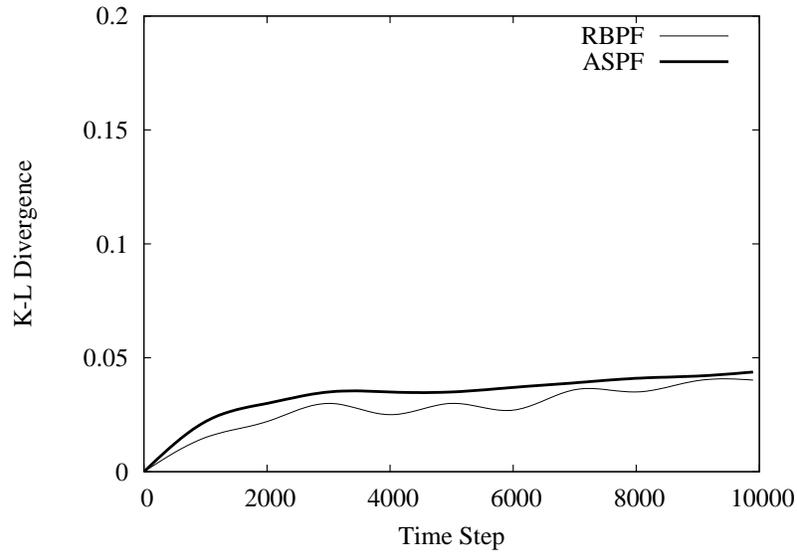

Figure 14:  Although ASPF (20,000 particles) does not require A1 or A2, it is nearly as accurate as RBPF (20,000 particles) when A1 and A2 hold and hence RBPF is applicable.

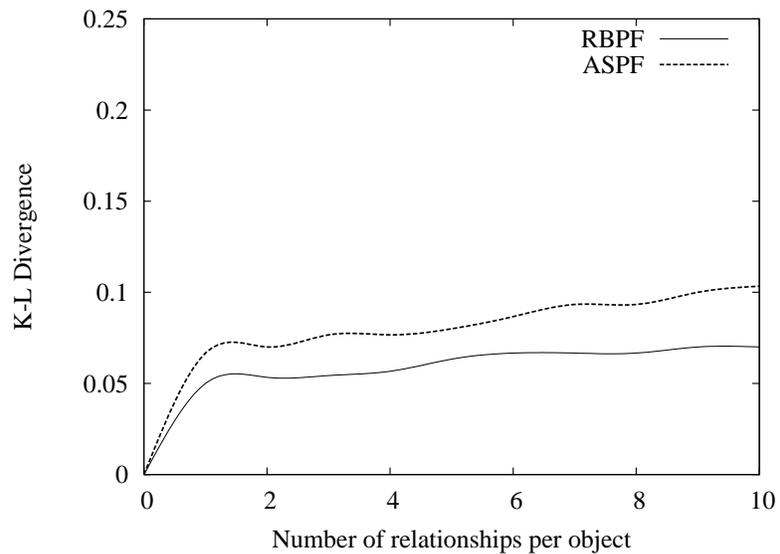

Figure 15:  The K-L divergence difference between ASPF and RBPF is small irrespective of the number of relations per object.

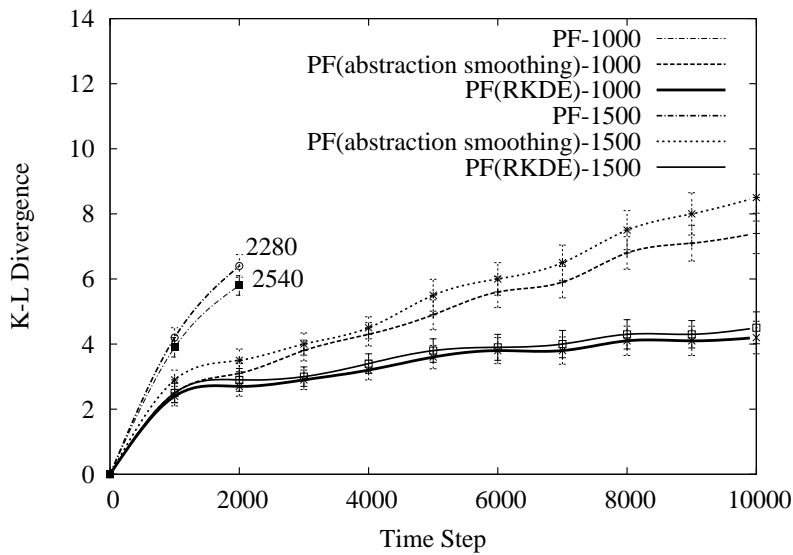

Figure 16: PF with relational kernel density estimation outperforms standard PF and PF with abstraction smoothing using independence assumptions. The experiment was run with 1000 and 1500 objects and $p_f = 1\%$, with our algorithms using 20,000 particles and standard PF using 100,000 particles.

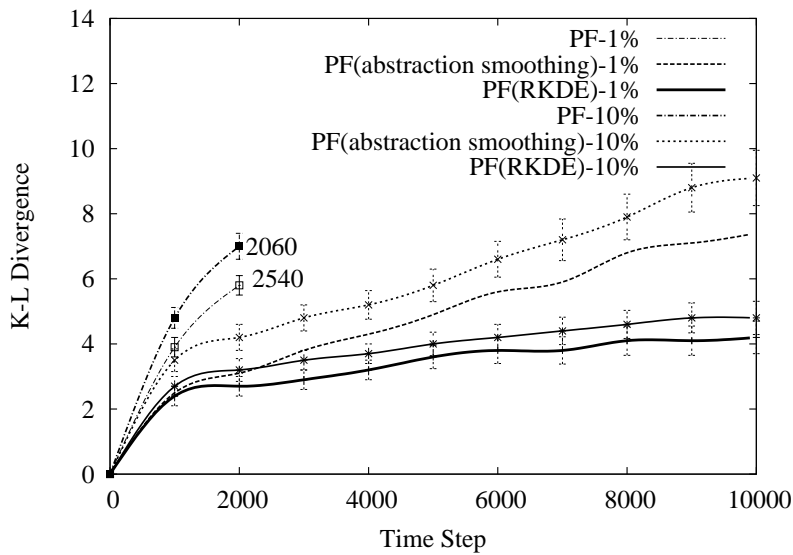

Figure 17: PF with relational kernel density estimation outperforms the other algorithms when predicting the joint distribution. The experiment was run with 1000 objects and fault probabilities of 1% and 10%.

## 7.5 Experimental Conclusions

The following are the conclusions that we can draw from the experiments:

- All of our algorithms (RBPF, PF with smoothing and RKDE) are much more accurate than standard PF for inference in RDBNs, using similar computational resources.

- RBPF is the best method when Assumption A1 holds, and scales up to a small maximum number of relations per object per predicate.

- PF with abstraction smoothing, unlike RBPF, is applicable in all scenarios to estimate the marginal distributions and is quite accurate.

- For estimating joint distributions, relational kernel density estimation outperforms PF with abstraction smoothing.

## 8. Related Work

In recent years, much research has focused on extending Bayesian networks to domains with relational structure. Approaches include stochastic logic programs (Muggleton, 1996; Cussens, 1999), probabilistic relational models (Friedman et al., 1999; Getoor, Friedman, Koller, & Taskar, 2001), Bayesian logic programs (Kersting & De Raedt, 2000) and Markov logic networks (Richardson & Domingos, 2004), among others. The relational Bayesian networks as we have defined in this paper are most closely related to the *recursive relational Bayesian networks* of Jaeger (1997). The main difference is that we specify the probabilistic dependencies using FOPTs whereas Jaeger uses the notion of combination functions (such as noisy-or) and equality constraints to define *probability formulae* over multisets.

However, there has been very limited work on extending these to temporal domains. Dynamic object-oriented Bayesian networks (DOOBNs) (Friedman, Koller, & Pfeffer, 1998) combine DBNs with OOBNs, a predecessor of PRMs. Unfortunately, no efficient inference methods were proposed for DOOBNs, and they have not been evaluated experimentally. Glesner and Koller (1995) proposed the idea of adding the power of first-order logic to DBNs. However, they only give procedures for constructing flexible DBNs out of first-order knowledge bases, and do not consider inference or learning procedures. Like DOOBNs, these were also not evaluated experimentally. Relational Markov models (RMMs) (Anderson et al., 2002) and logical hidden Markov models (LOHMMs) (Kersting & Raiko, 2005) are an extension of HMMs to first-order domains.

In our previous work, we introduced dynamic probabilistic relational domains (DPRMs) (Sanghai et al., 2003) which are an extension of PRMs. DPRMs can be viewed as a combination of PRMs and dynamic Bayesian networks. DPRMs are based on frame-based systems, which model the world in terms of classes, objects and their attributes. Objects are instances of classes, and each class has a set of propositional attributes and relational attributes (reference slots). The propositional attributes represent the properties of an object and the relational attributes model relationships between two objects. A DPRM specifies a probability distribution for each attribute of each class as a conditional probability table. The parents of an attribute are other attributes of the same class or attributes of related classes reached via some slot chain. A slot in a frame-based system performs the same function as a foreign key in a relational database. A slot chain can be viewed as a sequence of foreign keys enabling one to move from one table to another. The parents can be attributes from the current time slice or previous time slices.

DPRMs are related to RDBNs in the same way that frame-based systems are related to first-order logic. In Appendix A we prove that RDBNs subsume DPRMs, i.e., for every DPRM representing a probability distribution over a dynamic relational domain, there exists an RDBN which gives the same distribution. The proof is straightforward and it involves replacing attributes by predicates and conditional probability tables (CPTs) by first order probability trees (FOPTs). One of the important points to note is that all the inference algorithms described here are also applicable to DPRMs. There are also several advantages of using RDBNs instead of DPRMs:

- RDBNs generalize DPRMS by providing a more powerful language (first-order logic instead of frame-based systems).

- In DPRMs, the parents of an attribute (i.e., predicate) are obtained by traversing chains of reference slots, which correspond to conjunctive expressions, while in RDBNs parents can be obtained via any first-order logic constraints. However, because of the restrictions placed on DPRMs, learning them is potentially easier.

- Modeling n-ary relationships using DPRMs requires breaking them up into binary relationships (slots), which makes the task cumbersome. In general, the language of DPRMs is much harder to understand than that of RDBNs.

- In DPRMs the set of parents and the conditional model for each attribute are specified using one big table. In RDBNs, the parents and the conditional model are specified using FOPTs which can take advantage of context-specific independence to reduce space requirements and possibly speed up inference.

- In DPRMs, mutual exclusion between ground predicates is not modeled. For example, when modeling multi-valued slots, i.e., cases in which each object can be related to many other objects via the same slot, independence is assumed between the participating target objects.

- RDBNs have more scope for learning using ILP techniques such as first-order decision tree induction (Blockeel & De Raedt, 1998).

Particle filtering is currently a very active area of research (Doucet et al., 2001). The FastSLAM algorithm uses a tree structure to speed up RBPF with Gaussian variables (Montemerlo, Thrun, Koller, & Wegbreit, 2002). Koller and Fratkina (1998) used the particles at each step to induce a distribution over the DBN's states, and generated the next step's particles from this distribution. We tried this approach, but it led to poor results compared to using the distribution only to estimate the current state. Koller and Fratkina found that density trees outperformed Bayesian networks as the representation for the distribution. Consistent with these results, we tried Bayesian networks and found they were less accurate than abstraction trees (and also much slower).

An alternate method for efficient inference in DBNs that may also be useful in RDBNs was proposed by Boyen and Koller (1998) and combined with particle filtering by Ng et al (2002). These methods take advantage of the structure in the network, decomposing it into several nearly independent parts and performing inference separately on each of them. Efficient MCMC inference in relational probabilistic models has been studied by Pasula and Russell (2001). Their method uses a Metropolis-Hasting's step instead of the standard Gibbs step to sample relational variables, and is applicable only to a restricted form of distribution.

The use of abstractions has been studied extensively in AI (e.g., Koenig & Holte, 2002). Friedman et al (2000) used value abstractions to compute the likelihood. They defined *safe* and *cautious* abstractions with respect to a variable and such a concept can also be used in RDBNs to speed up computation. However, they did not consider using a hierarchy of abstractions and did not smooth over the abstractions. Verma et al (2003) used abstractions for particle filtering in small numeric DBNs, with a bias-variance criterion for choosing abstractions; this technique may be generalizable to RDBNs.

Downstream, RDBNs should be relevant to research on relational Markov decision processes (e.g., Boutilier, Reiter, & Price, 2001).

## 9. Conclusions and Future Work

This paper introduces relational dynamic Bayesian networks (RDBNs), a representation that handles time-changing phenomena, relational structure and uncertainty in a principled manner. We develop three approximate algorithms for doing efficient inference in RDBNs:

- Rao-Blackwellized particle filtering extends standard particle filtering by analytically computing the joint distribution of the complex predicates given sampled instances of the simple predicates samples. The method only applies when the complex predicates do not appear as parents of other predicates in the RDBN. When the number of relations per object is bounded by a small constant (e.g. 10), Rao-Blackwellization can be done efficiently and it greatly outperforms standard particle filtering.

- Particle filtering with abstraction-based smoothing uses abstraction lattices defined over the complex predicates to smooth the particle filter estimates. When computing marginal distributions, particle filtering with abstraction-based smoothing requires substantially fewer particles than standard particle filter and gives very accurate results.

- Relational kernel density estimation is an extension of particle filtering used to compute joint distributions by defining a kernel function over the ground state of the complex predicates. The relational kernel density estimation outperforms both standard particle filtering and particle filtering with abstraction-based smoothing when predicting joint distributions of the unknown predicates.

The above algorithms can be used in any relational stochastic process. They can also be applied to static relational domains and propositional domains. In the future we wish to combine smoothing and kernel density estimation with sampling algorithms like MCMC. Other directions for future work include handling continuous variables, learning RDBNs, using them as a basis for relational MDPs, and applying them to increasingly-complex real-world problems.

## Acknowledgements

This work was partly funded by NSF grant IIS-0307906, ONR grants N00014-02-1-0408 and N00014-02-1-0932, DARPA project CALO through SRI grant number 03-000225, a Sloan Fellowship to the second author, and an NSF CAREER Award to the second author.

## Appendix A: Assembly Domain

This section contains a list of the objects, properties(simple predicates), relations (complex predicates) and actions used in the experiments.

- Objects:

    - Plates (Shape, Surface, Temperature, Color, Size)

    - Brackets (Shape, Surface, Weight, Color)

    - Bolts (Hole_Type, Size, Weight)

- Relations:

    - Welded-To: (Plate, Plate), (Plate, Bracket)

    - Bolted-To: (Plate, Bolt), (Bracket, Bolt)

- Propositional Actions

    - Lathe (Object, Shape/Size, Value, t)

    - Paint (Object, Color, Value, t)

    - Polish (Object, Surface, Value, t)

    - Heat (Object, Temperature, Value, t)

    - Punch (Object, Hole_Type, Value, t)

- Relational Actions

    - Weld (Plate, Plate/Bracket, t)

    - Bolt (Bolt, Plate/Bracket, t)

- Fault models
  We now describe the fault model for the propositional and relational actions. These fault models in our system are defined using FOPTs. However, they turn out to be complex, particularly for relational actions. Hence, we describe them here using pseudo-code, where $f_p$ is the fault probability.

---

**Algorithm 2** Fault model for propositional actions

---

*PropAction (Object, Attribute, Value, t)*

Do one of the three actions below:

1. $Attribute(Object, Value, t) \leftarrow$ True
   $Attribute(Object, Value', t) \leftarrow$ False for all $Value' \neq Value$
   with probability $p = 1 - f_p$.

2. Leave the state unchanged
   with probability $p = f_p/2$

3. Choose $Value'$ with uniform distribution
   $Attribute(Object, Value', t) \leftarrow$ True
   $Attribute(Object, Value'', t) \leftarrow$ False for all $Value'' \neq Value'$
   with probability $p = f_p/2$

---

---

**Algorithm 3** Fault model for relational actions

---

*Weld (O1, O2, t)*

Do one of the five actions below:

1. *Welded-To (O1, O2, t)* ← True
   with probability $p = 1 - f_p$

2. Leave state unchanged
   with probability $p = 0.05 * f_p$

3. Choose a wrong object $O'$ with probability $p = 0.45 * f_p$ as follows

   - With $p \propto 1/2$ choose $O'$ with uniform distribution from the set of objects whose *color* and *shape* are the same as that of $O1$

   - With $p \propto 1/8$ choose $O'$ with uniform distribution from the set of objects whose *color* is the same as that of $O1$

   - With $p \propto 1/8$ choose $O'$ with uniform distribution from the set of objects whose *shape* is the same as that of $O1$

   - With $p \propto 1/32$ choose the object $O'$ uniformly from the set of all objects

   *Welded-To(O', O2, t)* ← True.

4. Choose a wrong object $O'$ to replace $O2$ with probability $p = 0.45 * f_p$, using the above procedure
   *Welded-To(O1, O', t)* ← True.

5. Choose wrong objects $O1'$ and $O2'$ with probability $p = 0.05 * f_p$ using the above procedure
   *Welded-To(O1', O2', t)* ← True.

*Bolt (O1, O2, t)*

Fault model is precisely the same as above except that the properties on which the choice of wrong object depends are *Size* for bolts and *Color* and *Surface* for plates and brackets.

---

## Appendix B: RDBNs as a Generalization of DPRMs

As discussed before, dynamic probabilistic relational models (DPRMs) can also be used to model uncertainty in a dynamic relational domain. However, they are an extension of PRMs, which are based on frame-based systems, and inherit the limitations of the latter. Here we show that RDBNs subsume DPRMs. In Section 8 we saw that the converse is not true.

**Proposition 3** *For each DPRM representing a probability distribution over dynamic relational domains there is an RDBN representing the same distribution.*

*Proof*: We will first convert the attribute and the reference slots in a DPRM to predicates in RDBNs, then add the corresponding edges to indicate the parents of a node, and finally prove that this does not lead to a cycle and the CPT can be converted to a FOPT.

Each slot $C.A$ in a DPRM corresponds to a predicate $A(x, v, t)$ where $A$ is the predicate name, $x$ represents the object, $v$ is the value of the slot and $t$ indicates the time slice. If $A$ is a simple attribute, then $v$ is a constant representing the value, otherwise $v$ is an object. If $C.A$ has a parent of the form $C.B$, then $A(x, y, t)$ has $B(x, y, t)$ as the parent. If $C.A$ has a parent of the form $\gamma(C.\tau.B)$ where $\tau$ is a slot chain and $\gamma$ an aggregation function, then all of the predicates corresponding to the slots in the slot chain are parents of $A(x, y, t)$.

We now show that if the initial DPRM is legal (i.e., without a cycle) then the RDBN obtained above is also legal. To prove this, we first consider the set of all certain slots in the DPRM (i.e., slots whose values are already known). In the RDBN, the predicates corresponding to these are certain and these predicates can be given higher priority than any of the other predicates. The relative ordering of the predicates themselves does not matter. For the rest of the predicates we define a relative ordering as follows: if any predicate appears as a parent of some other predicate, the parent predicate is given a higher priority.

We have to prove that the relative ordering defined above is consistent, i.e., if we consider the RDBN graph there is no cycle among the predicates. We do this by contradiction.

Assume that there is a cycle corresponding to predicates $R_1, \cdots R_k$, i.e., $R_i$ is a parent of $R_{i-1}$ and $R_1$ is a parent of $R_k$. It is easy to see that any predicate which has known values cannot appear in the cycle. If the cycle consists of only predicates which correspond to simple attributes, we can see that there will be a cycle among the attributes in the DPRM. Therefore, the cycle must involve predicates which correspond to reference slots (that are unknown). We will assume that all the predicates in the cycle correspond to reference slots and prove that the DPRM is illegal, leading to a contradiction. The case where some of the predicates in the cycle might correspond to simple attributes can be ruled out similarly. Let $C_i.\rho_i$ be the reference slot corresponding to the predicate $R_i$. Since $R_i$ is a parent of $R_{i-1}$ either $C_i.\rho_i$ is a parent of $C_{i-1}.\rho_{i-1}$ in the DPRM or $C_i.\rho_i$ appears in the slot chain $\tau$ such that $C_{i-1}.\tau.B$ is a parent of $C_{i-1}.\rho_{i-1}$. In the first case there is an edge in the DPRM from $C_i.\rho_i$ to $C_{i-1}.\rho_{i-1}$. In the second case, there is an edge from $C_i.\rho_i$ to $C_{i-1}.\rho_{i-1}$ if $C_i.\rho_i$ is unknown, which is the case here. Hence, we can see that a cycle among the predicates corresponds to a cycle among the reference slots in the DPRM. This implies that the DPRM is illegal, leading to a contradiction.

Finally, we have to prove that the CPT in a DPRM can be converted to a FOPT in an RDBN. We can first see that if in the DPRM $C.B$ is a parent of $C.A$, then we make $B(x, v, t)$ a parent of $A(x, v', t)$ for all values of $v$. However, we need to make sure that $A(x, v', t)$ is false if $A(x, v'', t)$ is true. To do this we define an (any) ordering on the constants, we make $A()$ a parent of itself, and we introduce the requisite dependencies, restricted to higher-priority groundings. Similarly, if $\gamma(C.\tau.B)$ is a parent of $C.A$, where $\gamma$ is an aggregation function, then in the FOPT we can have the expression $\gamma(v : \exists_{y_1,\cdots,y_{m+1}} R_1(x, y_1, t) \wedge \cdots \wedge R_i(y_i, y_{i+1}, t) \wedge \cdots B(y_{m+1}, v, t))$ where $\tau$ is the slot chain $\rho_1...\rho_m$ and $R_i$ is the predicate corresponding to the slot $\rho_i$. We can now easily see that for a node $A(x, v', t)$, any row of the corresponding CPT in the DPRM is equivalent to a first-order expression in the FOPT involving the expression above and tests on $B(x, v, t)$ and $A(x, v'', t)$. Hence the CPT can be converted to an equivalent FOPT. $\square$